\documentclass[acmlarge]{acmart}

\usepackage{amsmath,amsfonts} 
\usepackage[ruled,vlined]{algorithm2e}
\usepackage{graphicx}
\usepackage{textcomp}
\usepackage{xcolor}
\usepackage{tabularx}
\usepackage{caption}
\usepackage{subcaption}
\usepackage[inline]{enumitem}
\newcommand{\Ch}{\checkmark}
\newcommand{\X}{$\times$}
\usepackage{bm}
\usepackage{bbm}
\usepackage{url}
\usepackage{amsthm}
\usepackage{wrapfig}
\usepackage{multicol}
\usepackage{multirow}

\SetCommentSty{mycommfont}

\usepackage{xspace}
\newcommand{\Method}{SensorChat\xspace}
\newcommand{\MethodC}{SensorChat\textsubscript{C}\xspace}
\newcommand{\MethodE}{SensorChat\textsubscript{E}\xspace}
\newcommand{\Dataset}{SensorQA\xspace}
\newcommand{\citesensorqa}{\cite{sensorqa}\xspace}

\newcommand*{\revise}{\textcolor{black}}
\definecolor{myred}{RGB}{170,0,0} 
\definecolor{myblue}{RGB}{0,0,170} 
\definecolor{mygreen}{RGB}{0,100,0} 
\usepackage[most]{tcolorbox}

\newtcolorbox{mytextbox}[1][]{%
  sharp corners,
  enhanced,
  colback=white,
  height=2.6cm,
  attach title to upper,
  #1
}

\newtcolorbox{mytextbox2}[1][]{%
  sharp corners,
  enhanced,
  colback=white,
  height=5.2cm,
  attach title to upper,
  #1
}

\AtBeginDocument{%
  }

\setcopyright{acmlicensed}
\copyrightyear{2025}
\acmYear{2025}
\acmDOI{XXXXXXX.XXXXXXX}

\begin{document}

\title{\Method: Answering Qualitative and Quantitative Questions during Long-Term Multimodal Sensor Interactions}




\author{Xiaofan Yu}
\email{x1yu@ucsd.edu}
\orcid{0000-0002-9638-6184}
\affiliation{%
  \institution{University of California San Diego}
  \city{La Jolla}
  \state{California}
  \country{USA}
}

\author{Lanxiang Hu}
\email{lah003@ucsd.edu}
\orcid{0000-0003-0641-3677}
\affiliation{%
  \institution{University of California San Diego}
  \city{La Jolla}
  \state{California}
  \country{USA}
}

\author{Benjamin Reichman}
\email{bzr@gatech.edu}
\orcid{0009-0004-3854-7930}
\affiliation{%
  \institution{Georgia Institute of Technology}
  \city{Atlanta}
  \state{Georgia}
  \country{USA}
}

\author{Dylan Chu}
\email{dchu@ucsd.edu}
\orcid{0009-0001-5511-3286}
\affiliation{%
  \institution{University of California San Diego}
  \city{La Jolla}
  \state{California}
  \country{USA}
}

\author{Rushil Chandrupatla}
\email{ruchandrupatla@ucsd.edu}
\orcid{0009-0006-5447-8693}
\affiliation{%
  \institution{University of California San Diego}
  \city{La Jolla}
  \state{California}
  \country{USA}
}

\author{Xiyuan Zhang}
\email{xiyuanzh@ucsd.edu}
\orcid{0000-0002-8908-1307}
\affiliation{%
  \institution{University of California San Diego}
  \city{La Jolla}
  \state{California}
  \country{USA}
}

\author{Larry Heck}
\email{larryheck@gatech.edu}
\orcid{0000-0003-3358-6362}
\affiliation{%
  \institution{Georgia Institute of Technology}
  \city{Atlanta}
  \state{Georgia}
  \country{USA}
}

\author{Tajana \v{S}imuni\'{c} Rosing}
\email{tajana@ucsd.edu}
\orcid{0000-0002-6954-997X}
\affiliation{%
  \institution{University of California San Diego}
  \city{La Jolla}
  \state{California}
  \country{USA}
}

\renewcommand{\shortauthors}{}
\newcommand{\xiyuan}[1]{\textcolor{violet}{\textbf{Xiyuan:} #1}}
\newcommand{\lx}[1]{\textcolor{blue}{\textbf{Lanxiang:} #1}}

\begin{abstract}
\revise{Natural language interaction with sensing systems is crucial for addressing users’ personal concerns and providing health-related insights into their daily lives.
When a user asks a question, the system automatically analyzes the full history of sensor data, extracts relevant information, and generate an appropriate response. 
However, existing systems are limited to short-duration (e.g., one minute) or low-frequency (e.g., daily step count) sensor data. In addition, they struggle with quantitative questions that require precise numerical answers.}

\revise{In this work, we introduce \Method, the first end-to-end QA system designed for daily life monitoring using long-duration, high-frequency time series data. Given raw sensor signals spanning multiple days and a user-defined natural language question, \Method generates semantically meaningful responses that directly address users’ concerns. \Method effectively handles both quantitative questions that require numerical precision and qualitative questions that require high-level reasoning to infer subjective insights.}
To achieve this, \Method uses an innovative three-stage pipeline including question decomposition, sensor data query, and answer assembly.
\revise{The first and third stages leverage Large Language Models (LLMs) to interpret human queries and generate responses.
The intermediate querying stage extracts relevant information from the complete sensor data history, which is then combined with the original query in the final stage to produce accurate and meaningful answers.}
Real-world implementation demonstrate \Method's capability for real-time interactions on a cloud server while also being able to run entirely on edge platforms after quantization. Comprehensive QA evaluations show that \Method achieves up to \revise{93\%} higher answer accuracy than state-of-the-art systems on quantitative questions. Additionally, a user study with eight volunteers highlights \Method's effectiveness in answering qualitative and open-ended questions.
\end{abstract}



\settopmatter{printacmref=false}


\maketitle

\section{Introduction}
\label{sec:intro}





\revise{In recent years, the rapid growth of Internet of Things (IoT) devices has made them a ubiquitous part of daily life, from smartphones and smart home accessories to wearable technologies~\cite{choi2021smart}. These devices, equipped with multimodal on-device sensors, enable continuous monitoring of human activities and environmental conditions. The large volumes of long-term sensor data collected by these devices can provide a comprehensive view of users' daily routines and valuable insights into their health.
While traditional research has focused on building classification models for tasks like human activity recognition~\cite{ouyang2022cosmo, ouyang2023harmony, xu2023practically, xu2023mesen, cao2024mmclip, weng2024large}, recent work has highlighted the value of natural language interfaces for making sensor data more accessible and useful to humans~\cite{xing2021deepsqa, nie2022ai, yang2024drhouse, arakawa2024prism, ji2024mindguard, chen2024sensor2text}.
Question Answering (QA) systems, where users ask questions and receive responses from the sensor systems, offer a flexible and intuitive way to engage with sensor data.  Unlike passive lifelogs or visualizations~\cite{aizawa2004efficient,takahashi2013design,hwang2014lifelog,adiga2020daily,xu2024autolife}, QA enables users to actively express their information needs, making the system more responsive to their personalized concerns.}

\revise{
A useful QA system should be able to handle long-term sensor data and a wide range of practical queries. We envision users collecting sensor data through everyday mobile devices (like smartphones and smartwatches) equipped with multimodal sensors, such as IMUs, microphones, GPS, and compasses. These sensors operate continuously in the background, capturing rich streams of data that can be stored locally or sent to the cloud.
When users have specific questions or want to seek insights, they interact with the QA system, which automatically understands their intent, analyzes historical raw sensor data to extract relevant information, and generates meaningful answers. The QA system can run either on mobile devices or in the cloud, depending on resource needs and privacy requirements. To build a truly practical QA system, we focus on \textit{long-duration} sensor data that spans more than a day, enabling the analysis of extended or recurring behavioral patterns (e.g., daily routines or lifestyle trends) that are not observable in shorter time frames. In addition, the system should be capable of processing \textit{high-frequency} raw time series data to capture fine-grained activities, like gym workouts, which cannot be detected through sparse sensor readings such as daily step counts. On the query side, existing research for daily life monitoring has primarily focused on two broad categories: \textit{quantitative} and \textit{qualitative} questions~\cite{xing2021deepsqa,sensorqa,henrich2000questions}. Quantitative questions require precise, objective answers derived from sensor data~\cite{xing2021deepsqa,sensorqa}, often involving content such as true/false, location, activity, count, or time duration (e.g.,  ``\textit{How long did I exercise yesterday?}''). In contrast, qualitative questions involve interpreting sensor data to infer high-level, subjective insights~\cite{henrich2000questions}. Qualitative questions can be vague, open-ended and often require external knowledge. For example, answering ``\textit{Did I exercise enough for a young adult?}'' requires understanding of the standard activity levels. 
We recognize that addressing both categories of questions is important in building a useful QA system.
To clarify our problem scope, we summarize the key concepts in Table~\ref{tbl:concepts}.}

\revise{In this paper, we explore the core research question: \textbf{How can we design a practical QA system that delivers accurate and meaningful answers to both quantitative and qualitative user queries, based on long-duration, high-frequency sensor data?} We recognize three key challenges for this problem:}

\begin{enumerate}
    \item \revise{\textbf{Processing long-duration and high-frequency sensor data} is inherently challenging due to the massive volume of raw data generated. For instance, a continuously running IMU sensor with 9 axes at a 40Hz sampling rate can produce roughly 125MB of data per day. In practice, mobile devices are equipped with a wide array of sensors beyond IMUs, further increasing the complexity and size of the data. Extracting relevant information from such vast, multimodal time series data becomes a difficult task that involves pattern recognition and temporal query retrieval.}
    \item \revise{\textbf{Effectively fusing variable duration of sensor data with textual questions} poses a significant challenge to answer certain questions. For instance, a specific quantitative question like "\textit{What was I doing at 10:00 AM yesterday?}" requires the system to precisely locate and interpret sensor data from a specific timestamp. In contrast, a high-level qualitative question such as "\textit{How would you rate my lifestyle?}" demands a broader understanding of patterns across extended periods. Effectively addressing these diverse queries requires flexible reasoning across multiple time scales, along with the ability to align temporal sensor data with natural language queries.}
    \item \revise{\textbf{Generating accurate and semantically meaningful answers} makes QA fundamentally more challenging than traditional classification or journaling tasks. In contrast to conventional human activity recognition~\cite{ouyang2022cosmo, ouyang2023harmony, xu2023practically, xu2023mesen, cao2024mmclip, weng2024large}, which typically involves selecting from 10–20 predefined activity classes, QA requires choosing the next appropriate token from a vocabulary of over 32,000 possibilities at each step~\cite{touvron2023llama}. Unlike passive journaling systems~\cite{aizawa2004efficient,takahashi2013design,hwang2014lifelog,adiga2020daily,xu2024autolife} that simply log events, QA systems need to understand the user's intent and deliver relevant, satisfying responses. The response needs to be a numerically precise answer to a quantitative question or a thoughtful, semantically rich interpretation for a qualitative question. This high level of complexity demands deeper reasoning, language understanding, and tight alignment between sensor data and user queries.}
\end{enumerate}

\begin{figure}[t]
  \begin{subfigure}[t]{0.38\textwidth}
        \centering
        \includegraphics[width=\textwidth]{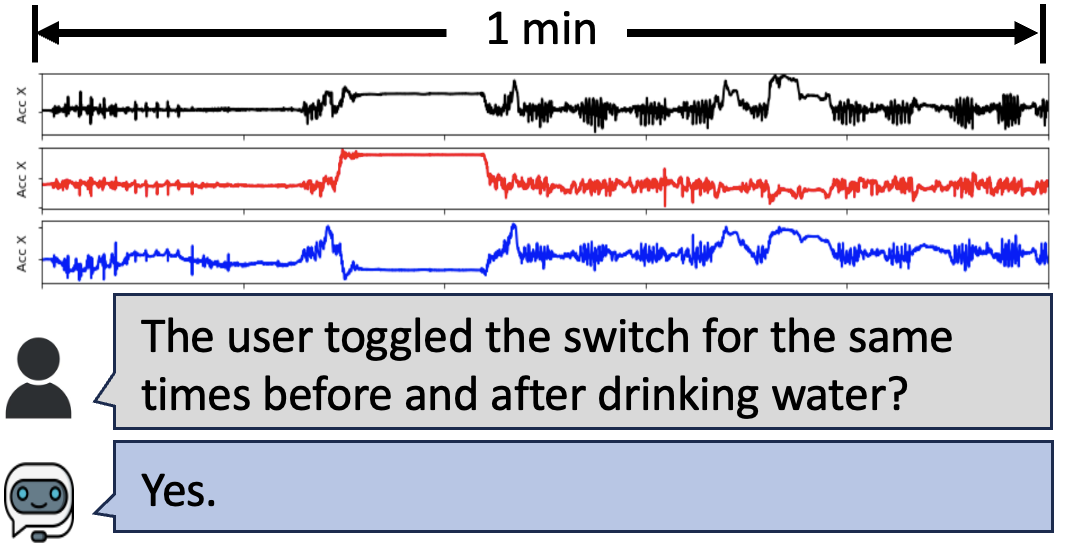}
        \vspace{-6mm}
        \caption{Example QA in DeepSQA~\cite{xing2021deepsqa} for human activities. \revise{The sensor duration is limited to 1 minute per conversation.}}
        \label{fig:sensor2text_example}
    \end{subfigure} \hspace{0.05\textwidth} 
    \begin{subfigure}[t]{0.45\textwidth}
        \centering
        \includegraphics[width=\textwidth]{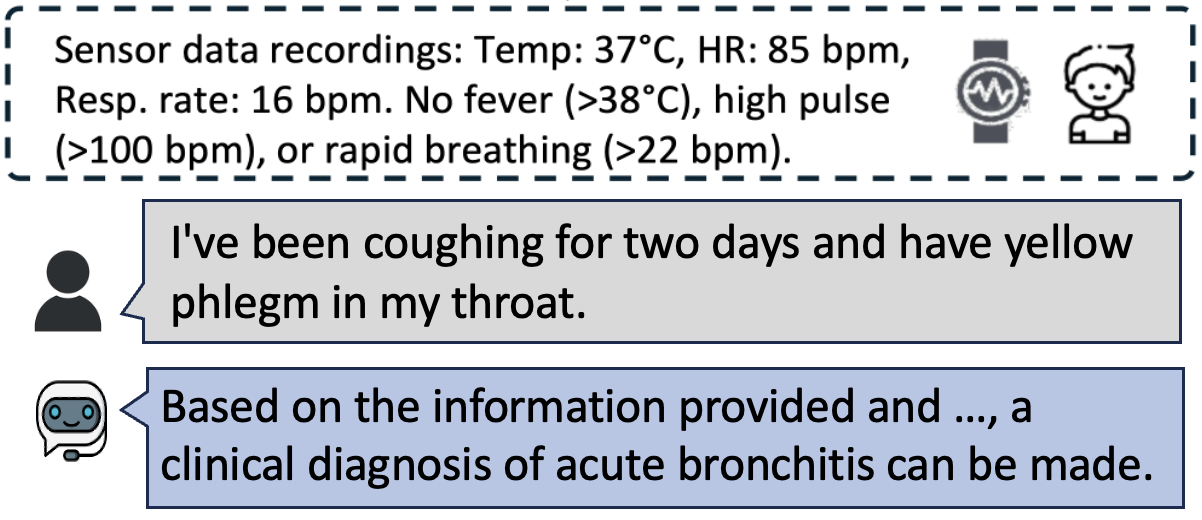}
        \vspace{-6mm}
        \caption{Example QA in DrHouse~\cite{yang2024drhouse} for clinical diagnosis. \revise{The sensor data is limited to sparse sensor readings thus may overlook fine-grained activities.}}
        \label{fig:drhouse_example}
    \end{subfigure}

    \begin{subfigure}[t]{0.92\textwidth}
        \centering
        \includegraphics[width=\textwidth]{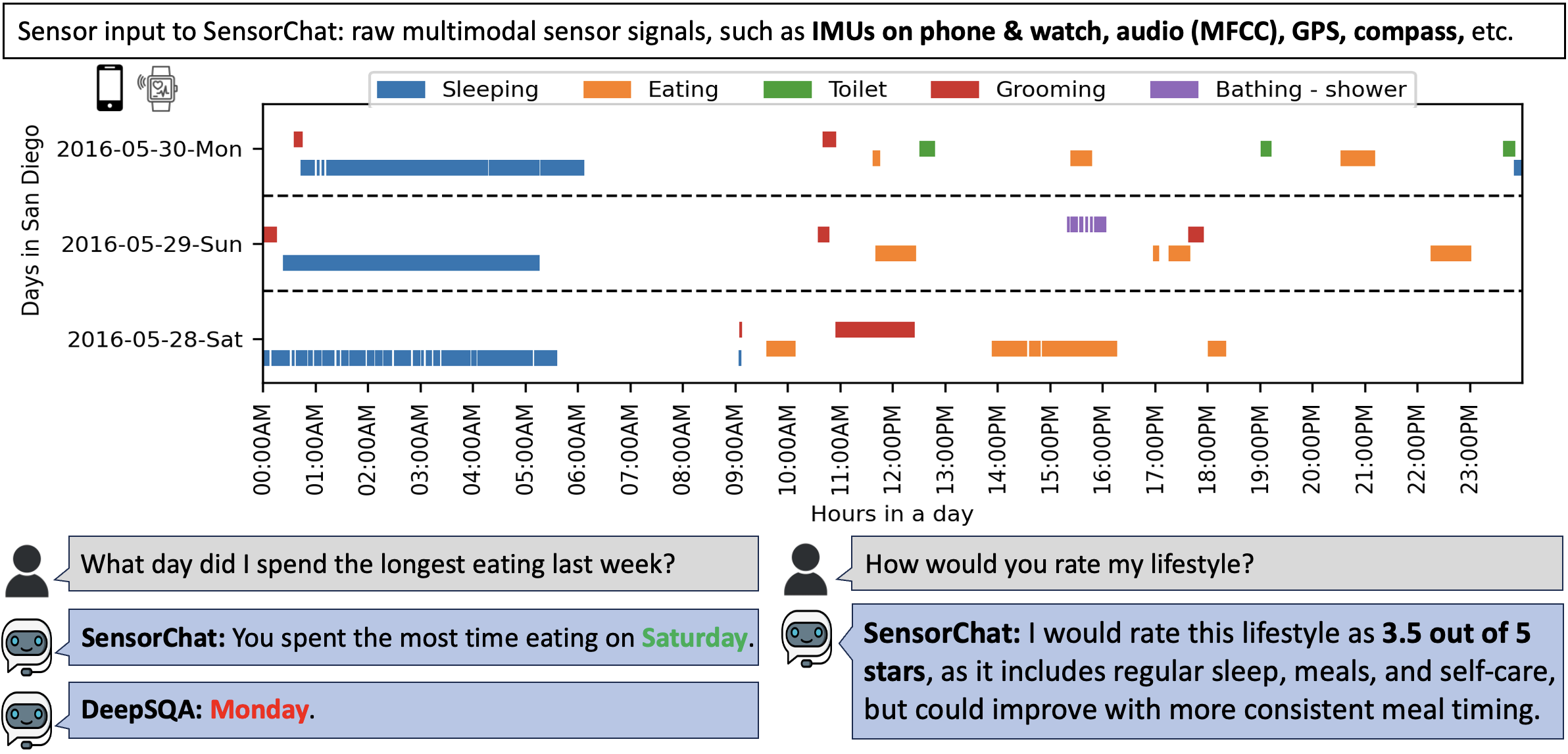}
        \vspace{-6mm}
        \caption{Example QA in \textbf{SensorChat (our system)}. \revise{In contrast to DeepSQA~\cite{xing2021deepsqa} and DrHouse~\cite{yang2024drhouse}, SensorChat handles long-duration and high-frequency raw sensor signals to accurately answer quantitative questions (left) and offer insightful qualitative analyses of well-being (right). DeepSQA~\cite{xing2021deepsqa}, when directly adapted to our sensor input, produces incorrect answers. The correct answer is in green while the incorrect one is in red.}}
        \label{fig:sensorchat_example}
    \end{subfigure}
    \vspace{-4mm}
    \caption{A visual comparison existing QA systems for sensor data and \Method.}
    \label{fig:moti_intro}
    \vspace{-6mm}
\end{figure}

\begin{table}[t]
\footnotesize
\caption{\revise{Key concepts that defines the problem scope of this paper. These concepts are motivated by practical user needs.}}
\label{tbl:concepts}
\vspace{-4mm}
\begin{center}
\begin{tabular}{p{2.8cm}|p{11cm}} 
\toprule
\textbf{Concept} & \textbf{Meaning} \\ \midrule
Long-duration data & Sensor data spans more than a full day of user activity \\
High-frequency data & Collected from high-frequency sensors (e.g., IMU raw signals), as opposed to sparse data like daily step counts or statistics (e.g., mean value) \\
Quantitative questions & Require precise and objective answers based on sensor data, such as true/false, location, activity, count, or time duration \\
Qualitative questions & Require high-level reasoning to infer subjective insights, such as work-life balance and social interactions, possibly involving external knowledge \\
\bottomrule
\end{tabular}
\end{center}
\vspace{-4mm}
\end{table}
\revise{Existing research falls short in addressing all above challenges. Current QA systems are limited to short sensor durations~\cite{xing2021deepsqa,chen2024sensor2text,arakawa2024prism,moon-etal-2023-imu2clip,han2024onellm} (e.g., one minute, as shown in Fig.~\ref{fig:sensor2text_example}) or low-frequency readings~\cite{englhardt2024classification,kim2024health,yang2024drhouse} (e.g., daily step counts and temperature, as shown in Fig.~\ref{fig:drhouse_example}). In Section~\ref{sec:motivation}, we demonstrate that directly applying these systems to long-duration, high-frequency sensor data results in poor answer quality due to the increased data volume and complexity.
Although recent advances in Large Language Models (LLMs) and multimodal LLMs offer new ways to understand raw sensor data, current approaches are ineffective for handling long-duration, high-frequency sensor data in our setting.
These approaches typically involve converting raw sensor signals into text~\cite{ji2024hargpt,xu2024penetrative,hota2024evaluating,liu2023large}, visualizing them as images~\cite{yoon2024my}, or encoding them into embeddings before feeding them into an LLM~\cite{moon2023anymal,han2024onellm,mo2024iot,chen2024sensor2text}. However, as detailed in Appendix~\ref{appendix:input-format}, uncompressed text or embeddings often exceed the token limit of LLMs (e.g., 8,092 tokens for GPT-4~\cite{gpt-4}), while compressing high-frequency sensor readings over a day into a single image results in loss of interpretability. To the best of our knowledge, no existing system effectively addresses all three challenges in our setting.
}

In this paper, we introduce \Method, the \textit{first} end-to-end system designed for answering both qualitative and quantitative queries in long-term sensor monitoring, using multimodal and high-frequency time series sensors.
In contrast to prior QA systems~\cite{xing2021deepsqa,chen2024sensor2text,arakawa2024prism,moon-etal-2023-imu2clip,han2024onellm,englhardt2024classification,kim2024health,yang2024drhouse}, \Method is capable of processing long-duration and high-frequency sensor data spanning weeks or even months. \revise{\Method generates accurate answers to quantitative questions and semantically meaningful answers to qualitative questions, all based on the extracted sensor information. Figure~\ref{fig:sensorchat_example} illustrates two QA examples using \Method. The input consists of raw multimodal sensor data collected over three days from a smartphone and a smartwatch. For clarity, we visualize the inferred user activity labels, with the x-axis showing time of day (00:00 AM to 11:59 PM) and the y-axis indicating separate days. Different activities are represented by color-coded bars.
When asked a quantitative query like ``\textit{What day did I spend the longest eating last week?}'', \Method automatically understands the question and searches the full sensor history to return a precise answer.
\Method can also handle open-ended qualitative questions, such as queries ``\textit{How would you rate my lifestyle?}'' by reasoning on top of a detailed analysis of daily routines.}

To achieve the goal, \Method introduces a novel three-stage design including question decomposition, sensor data query and answer assembly.
\revise{The key design philosophy of \Method is to use LLMs in the first and third stages to correctly understand user questions and generate natural language answers, while inserting an explicit query stage in between for accurate sensor data processing and effective sensor-question fusion. \Method tackles the first challenge of processing long-duration, high-frequency sensor data by pretraining a sensor encoder and a text encoder using a novel contrastive loss. This pretraining enables the model to associate raw sensor signals with semantically meaningful text embeddings. Then, in the second stage, \Method uses a custom query procedure to identify relevant sensor segments from the full monitoring history based on the textual input, bridging sensor data and natural language. Lastly, to ensure high-quality responses, \Method uses in-context learning~\cite{shao2023prompting} and Chain-of-Thought (CoT) prompting~\cite{chuCoTReasoningSurvey2024} during question decomposition to correctly interpret user intent. In the answer assembly stage, \Method applies Low Rank Adaptation (LoRA) finetuning~\cite{hu2021lora} to align answers with user preferences, based on a state-of-the-art dataset~\citesensorqa. The three stages of \Method work collaboratively to deliver accurate and meaningful responses to both quantitative and qualitative questions over long-term, high-frequency multimodal sensor data.}

We implement \Method in two system variants to accommodate various user needs: \MethodC (for cloud deployment) and \MethodE (for edge deployment). \MethodC uses OpenAI’s GPT APIs~\cite{gpt-3.5,gpt-4} to interpret user queries, offering real-time QA interactions on a server with an NVIDIA A100 GPU~\cite{a100}. On the other hand, \MethodE uses quantized LLMs and runs entirely on an edge device, preserving user privacy albeit at the cost of higher latency. We test \MethodE on an NVIDIA Jetson Orin NX module~\cite{jetsonorin}.
\revise{We evaluate \Method through two sets of experiments to test different capabilities of the system. First, we test \Method on \Dataset~\citesensorqa, a state-of-the-art QA dataset focused on quantitative questions. Next, we conduct a real-world user study with eight volunteers, approved by the Institutional Review Board (IRB), to assess \Method’s performance on both qualitative and quantitative questions. This study also tests \Method's generalizability to personalized mobile devices.} 

In summary, the major contributions of our work are:
\begin{itemize}[nolistsep]
    \item We propose \Method, the \textit{first} end-to-end system for answering qualitative and quantitative questions in long-term monitoring using high-frequency time series sensors. \Method is designed with a novel three stage pipeline, including question decomposition, sensor data query and answer assembly.
    \item We introduce a novel contrastive sensor-text pretraining loss for training the sensor and text encoders offline. \revise{We also design a custom query procedure to ensure accurate sensor embedding queries given arbitrary text during the online question-answering.}
    \item We enhance the LLMs in \Method through in-context learning, CoT and fine-tuning, incorporating carefully designed prompts to improve the quality of question decomposition and answer assembly.
    \item Evaluations on a \revise{quantitative} QA dataset show that \Method improves answer accuracy by up to \revise{93\%} compared to the best performing state-of-the-art QA systems. A real-world user study confirms that \Method provides meaningful and satisfying answers to users' questions, \revise{including both quantitative and qualitative ones,} while generalizing to personal mobile devices.
    \item \revise{We implement \Method in two system variants, \MethodC and \MethodE. \MethodC enables real-time user interactions from a cloud server, while \MethodE operates fully on local edge devices with reasonable answer latency.}
\end{itemize}

\section{Related Work}
\label{sec:related-work}

\begin{table}
\footnotesize
\caption{\revise{Comparison of \Method with existing sensor-based QA systems. \Method is the first QA system to provide accurate, meaningful answers based on long-duration, high-frequency time series sensor data.}}
\label{tbl:related_works}
\vspace{-4mm}
\begin{center}
\begin{tabular}{ccc} 
\toprule
\textbf{Existing QA systems} & \textbf{Long-duration} & \textbf{High-frequency} \\
& \textbf{sensor data} & \textbf{time series sensors} \\
\midrule
DeepSQA~\cite{xing2021deepsqa}, Sensor2Text~\cite{chen2024sensor2text}, PrISM-Q\&A~\cite{arakawa2024prism}, OneLLM~\cite{han2024onellm} & \X & \Ch \\
Englhardt \textit{et al.}~\cite{englhardt2024classification}, Health-LLM~\cite{kim2024health}, DrHouse~\cite{yang2024drhouse}& \Ch & \X \\ \midrule
\textbf{\Method (this work)} & \textbf{\Ch} & \textbf{\Ch} \\
\bottomrule
\end{tabular}
\end{center}
\vspace{-4mm}
\end{table}

\textbf{Question Answering using Sensor Data}
\revise{In the sensing domain, prior QA systems can be broadly categorized based on the type and duration of sensor data used, as summarized in Table~\ref{tbl:related_works}. One line of research focuses on understanding users' current activities using real-time sensor readings, as shown in Fig.~\ref{fig:sensor2text_example}. For each QA instance, DeepSQA~\cite{xing2021deepsqa} processes one-minute IMU signals, Sensor2Text~\cite{chen2024sensor2text} uses 16-second body tracking data, PrISM-Q\&A~\cite{arakawa2024prism} analyzes short-segment smartwatch signals for procedural guidance via a voice assistant, OneLLM~\cite{han2024onellm} allows 10-second IMU signals as one of the eight modailities. All of these works are limited to short-duration sensor data, whereas directly applying these methods to long-duration sensor data in our setting leads to ineffective fusion between sensor and text, as we detail in Sec.~\ref{sec:motivation}. On the contrary, \Method supports understanding and reasoning over long-term data spanning more than a day.
The second line of research uses long-term but sparse sensor data to support health-related queries and clinical assessments. Englhardt \textit{et al.}~\cite{englhardt2024classification}, Health-LLM~\cite{kim2024health}, and DrHouse~\cite{yang2024drhouse} converted physiological signals (e.g., heart rate, step counts) into text prompts for LLMs, as exemplified in Fig.~\ref{fig:drhouse_example}. However, these methods rely on low-frequency data, thus cannot capture fine-grained activities like gym sessions. In contrast, \Method processes raw, high-frequency sensor signals, capturing detailed user schedules and activity durations.}

\revise{\textbf{Multimodal LLMs using Sensor Data} The rise of LLMs has sparked growing interest in integrating sensor data into LLMs for high-level reasoning. HARGPT~\cite{ji2024hargpt}, Penetrative AI~\cite{xu2024penetrative}, Hota \textit{et al.}~\cite{hota2024evaluating}, and Liu \textit{et al.}~\cite{liu2023large} are among the first works to convert raw sensor signals into text prompts for LLMs. By-My-Eyes~\cite{yoon2024my} selects visualization tools to present sensor data as visual prompts. Other efforts in multimodal LLMs~\cite{moon2023anymal,han2023onellm} train adapters to convert raw sensor signals into embeddings, which are then used for finetuning LLMs. However, these methods struggle when tasked with interpreting the long-duration, high-frequency data in our scenario, as discussed in Appendix~\ref{appendix:input-format}. More recent approaches like IMU2CLIP~\cite{moon-etal-2023-imu2clip} and LLMSense~\cite{ouyang2024llmsense} use classifiers to map raw signals into narrative logs for LLM input, but still suffer from LLMs' limitations in handling long context~\cite{li2024long,gu2023mamba}, i.e., input with more than 2K tokens. In contrast, \Method overcomes this limitation by incorporating a dedicated sensor data query stage, enabling it to handle long-term sensor queries effectively.}
\section{Background and Motivation}
\label{sec:ackground-n-motivation}

In this section, we formally define the problem we address and present a motivating study highlighting the limitations of state-of-the-art QA techniques.


\subsection{Problem Statement}
\label{sec:problem-statement}


\revise{We formulate the problem as generating an appropriate answer based on a given question and sensor input. In the $i$th interaction round, the input and output are a pair of natural language question and answer, denoted as $(q^i, a^i)$ for $i \in {1, ..., N}$, where $N$ is the total number of QA interactions. The sensor input consists of the full history of raw sensor signals $\mathbf{x} \in \mathbb{R}^{T \times d}$, where $T$ is the number of recorded time steps and $d$ is the number of multimodal sensors. We specifically target long-duration, high-frequency sensor data (e.g., raw IMU signals sampled at 40Hz over more than a day), resulting in extremely large $T$.}
\revise{Formally, our objective is formed as training an autoregressive model, $\omega$, to predict the next token in the answer sequence:}
\begin{equation}
\max_\omega \sum_{i=1}^{N} \sum_{j=1}^{m} \log p_\omega \left ( a^i_j \: \mathlarger{|} \: \mathbf{x}, q^i, a^i_{1:j-1}\right ),\label{eq:problem}
\end{equation}
\revise{where $i$ iterates through all QA pairs, $j$ iterates through all tokens in the answer and $m$ is the maximal length for answers. Given a QA dataset $(q^i, a^i)$, $\omega$ is trained to maximize the likelihood of the correct next token. During inference, $\omega$ is fixed, and the model selects the next token with the highest predicted probability.}

\revise{\textbf{Question and Answer Types}
We target at two broad categories of user queries $q^i$: quantitative and qualitative questions, as commonly identified in prior work~\cite{xing2021deepsqa,sensorqa,henrich2000questions}. For quantitative questions, we mainly focus on the six question types and seven answer types defined by the SensorQA dataset~\cite{sensorqa}, a state-of-the-art quantitative QA benchmark for daily-life monitoring. Table~\ref{tab:sensorqa_profile} shows the detailed question and answer types. SensorQA is chosen for its realism, as its questions were written by human annotators simulating interactions with a smart sensor system.
Due to the lack of a benchmark for qualitative QA using long-term sensor data, we focus on general topics that affect quality of life, such as work–life balance and social interactions~\cite{henrich2000questions}. We evaluate \Method on qualitative questions through a real-user study in Sec.~\ref{sec:deployment}.
}

\subsection{Limitations of State-of-the-Art (SOTA) Methods}
\label{sec:motivation}
With a concrete problem setup, we conduct a motivating study to understand the limitations of state-of-the-art systems on this problem.
We then discuss the takeaways which inspire the design of \Method.


\textbf{Study Setup}
We select the \Dataset dataset\footnote{\url{https://github.com/benjamin-reichman/SensorQA}}~\citesensorqa because it has the best alignment with QA scenarios using long-duration, high-frequency multimodal sensor data. \Dataset has raw sensor measurements collected using smartphone and smartwatch sensors, including IMUs, audio (MFCC features), and phone state sensors (compass, GPS location, Wi-Fi, light, etc.). It contains data from 60 subjects and has more than 300K minutes of data, annotated with 51 \revise{context labels} after cleaning.
In total, \Dataset~\citesensorqa contains 5.6K quantitative QA pairs reflecting practical human interests in daily life monitoring, including approximately 3K questions on a single day and 2.6K for longer durations spanning weeks or months. We randomly split the data in \Dataset~\citesensorqa into an 80/20 train-test set. \revise{Raw sensor data is split into time windows of 20 seconds for models to process.}

\begin{table}[t]
\begin{subtable}[b]{0.6\textwidth}
\footnotesize
\centering
\begin{tabular}{c p{6.0cm}} 
\toprule
\textbf{Question Categories} & \textbf{Example Questions} \\
\midrule
Time Compare & Did I spend more time sitting or standing? \\
Day Query & On which day did I spend the most time at home? \\
Time Query & How long was I in class and at school?\\
Counting & How often did I groom? \\
Existence & Did I have a meeting on Wednesday? \\
Action Query & What did I do after I left home on Tuesday? \\
\bottomrule
\end{tabular}
\caption{Question categories.}
\label{tab:question_profile}
\end{subtable}
\hspace{0.02\textwidth}
\begin{subtable}[b]{0.35\textwidth}
\centering
\footnotesize
\begin{tabular}{c p{2.6cm}} 
\toprule
\textbf{Answer Categogies} & \textbf{Example Answers} \\
\midrule
Action & Doing computer work \\
Day/Days & Last Friday \\
Existence & Yes/No \\
Time Length & 40 Minutes \\
Location & At school \\
Count & Three times \\
Timestamp & Around 11:00 am \\
\bottomrule
\end{tabular}
\caption{Answer categories.}
\label{tab:answer_profile}
\end{subtable}
\vspace{-3mm}
\caption{\revise{Quantitative Q\&A categories that we focus on in this paper, as identified in the \Dataset dataset~\cite{sensorqa}.} The shorted version of the answers are presented for simplicity.}
\label{tab:sensorqa_profile}
\vspace{-7mm}
\end{table}

%



\revise{We train and evaluate three methods representing SOTA solutions using neural network and LLM-based models:
\textbf{(1) \textbf{LLaMA2-7B}~\cite{llama2}}, as a reference, is the LLM baseline that only considers the question without sensor data.
\textbf{(2) DeepSQA~\cite{xing2021deepsqa}} represents the SOTA CNN-LSTM based model for fusing the question and time series sensor data. While DeepSQA was originally designed for short-duration sensor data, we adapt it by using the full history of IMU signals as sensor-side input.
\textbf{(3) \textbf{LLMSense~\cite{ouyang2024llmsense}}} is the SOTA LLM-based method for reasoning over sensor data. It converts neural network classifications into a narrative log including all timestamps, and then feeds this log and the user question into an LLM to generate an answer. Detailed implementations of these methods are discussed in Appendix~\ref{appendix:baseline-implementation-details}.}
\revise{We focus on answer accuracy for quantitative questions in \Dataset. 
An answer is considered correct if it contains the key phrases from the ground-truth answer provided by \Dataset. Otherwise, it is marked as incorrect. For example, in Fig.~\ref{fig:sensorchat_example}, if the ground-truth answer is ``\textit{You spent the longest eating time on Saturday,}'' then an answer that includes "\textit{Saturday}", such as \Method's answer, is correct. An answer like "\textit{Monday}" is incorrect.}

\begin{figure*}[t]
\begin{center}
\begin{tabular}{cc}
    \multicolumn{2}{c}{\includegraphics[width=0.5\textwidth]{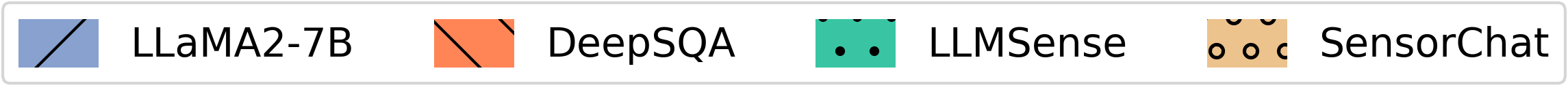}} \vspace{-1mm} \\
    \vspace{-1mm}
    \includegraphics[width=0.35\textwidth, height=2.5cm]{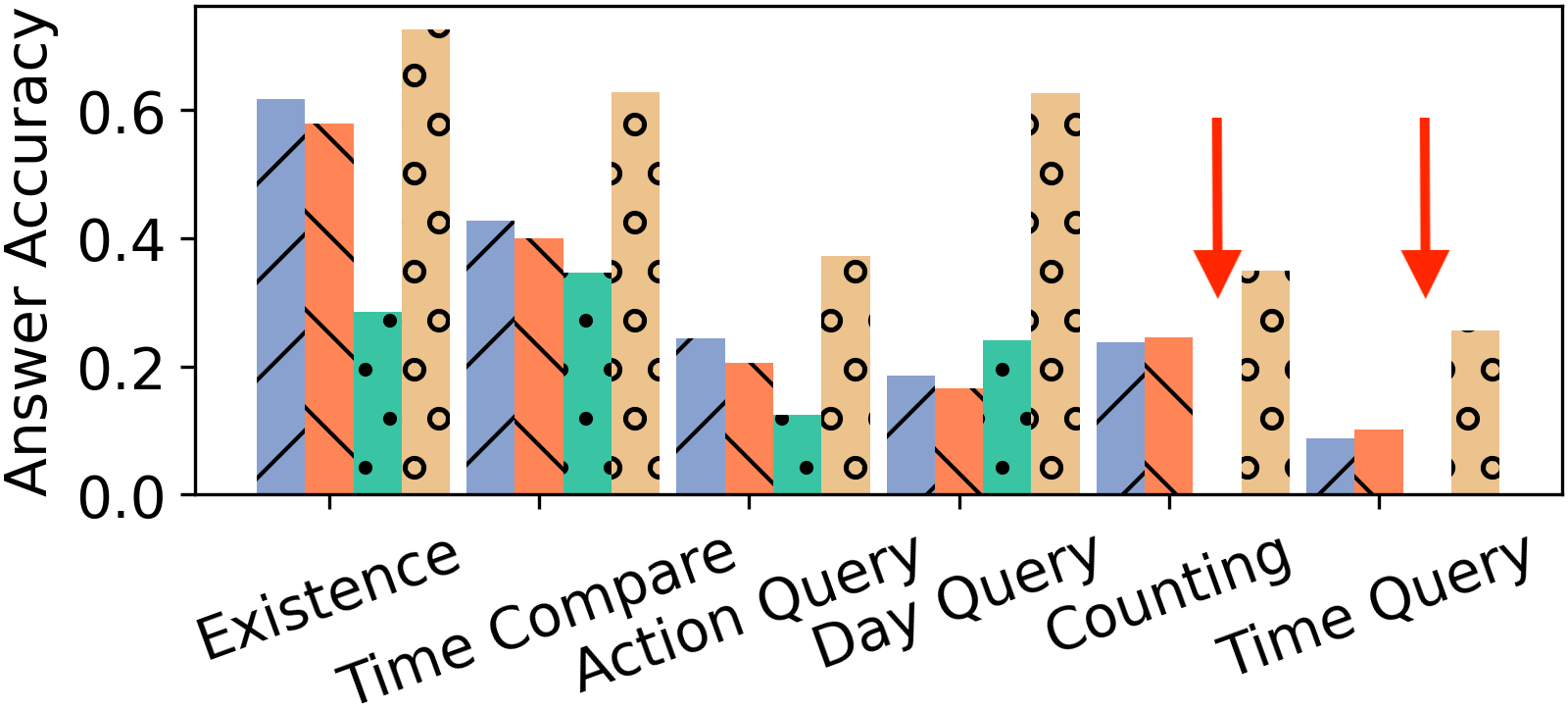} &  \includegraphics[width=0.35\textwidth, height=2.5cm]{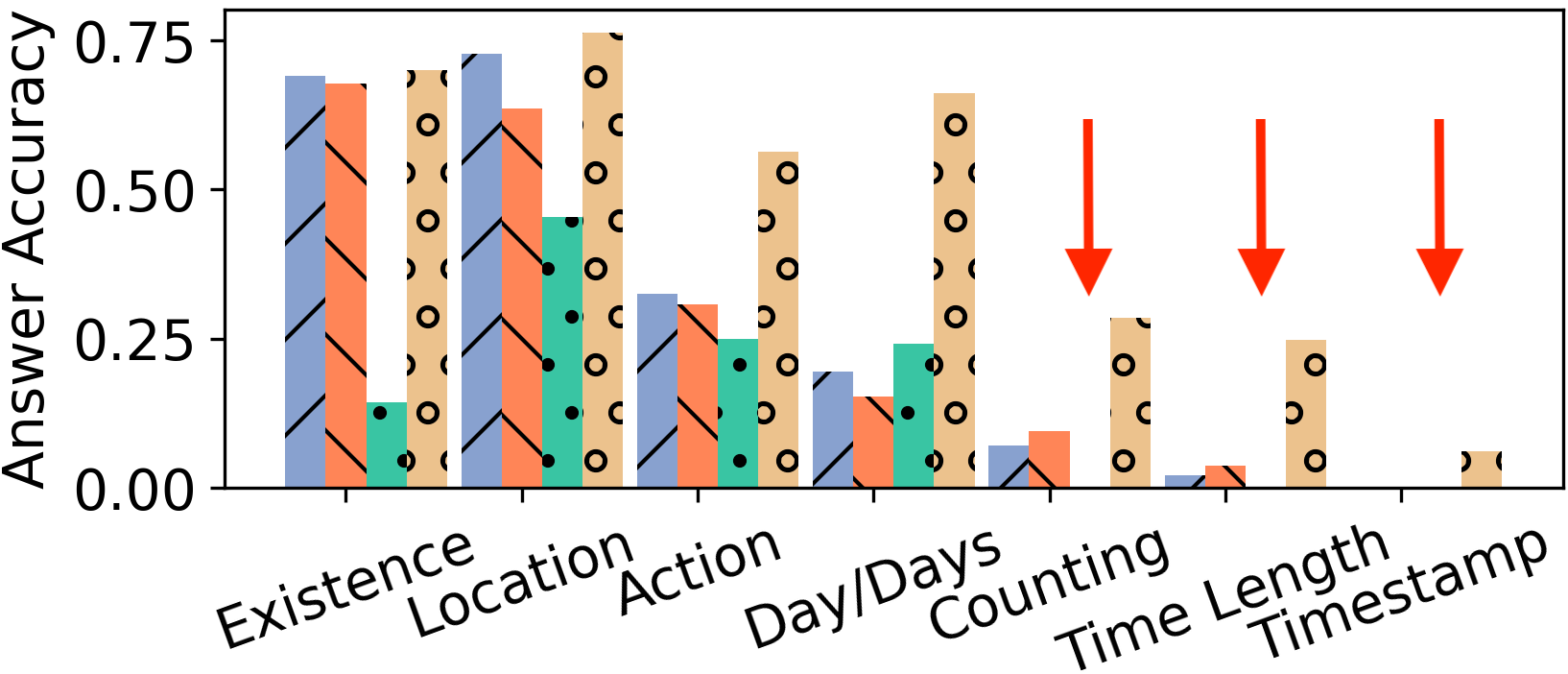}  \\
    (a) {\small \revise{Question categories}} & (b) {\small \revise{Answer categories}}  \\ 
\end{tabular}
\vspace{-4mm}
\caption{\revise{Answer accuracy of state-of-the-art methods and \Method on SensorQA~\citesensorqa. Categories are arranged from easier questions/answers on the left to more challenging questions/answers on the right. Red arrows highlight the question/answer types that existing works struggle with.}}
\label{fig:variants}
\end{center}
\vspace{-6mm}
\end{figure*}

\revise{Fig.~\ref{fig:variants} shows the accuracy results of SOTA methods by question category (left) and answer category (right).
The key observation is that all SOTA baselines deliver low answer accuracy. LLaMA2-7B~\cite{llama2}, DeepSQA~\cite{xing2021deepsqa}, and LLMSense~\cite{ouyang2024llmsense} achieve average accuracies of 0.26, 0.27, and 0.23, respectively. Notably, DeepSQA and LLMSense—both of which attempt to integrate sensor data with the question—perform comparably to or worse than LLaMA2-7B, which does not incorporate any sensor input. This suggests that existing approaches struggle to handle the complexity introduced by long-duration, high-frequency sensor data. In particular, DeepSQA’s performance suffers when directly fed with more than a day’s worth of IMU signals. The increased input size and complexity significantly complicate the fusion process, overwhelming its relatively simple model architecture. On the other end, LLMSense performs poorly on questions requiring precise numerical reasoning, such as counting or time queries (highlighted by red arrows in Figure~\ref{fig:variants}). This is due to LLMSense's reliance on LLMs to reason over long narrative logs, which contains up to 1,440 lines for a single day. Prior studies have shown that LLMs often struggle with long-context numerical tasks and are prone to hallucinations in such settings~\cite{huang2023survey,gu2023mamba,li2024long}.}

\revise{\textbf{In summary, directly applying existing QA methods to long-duration, high-frequency sensor data leads to poor performance, particularly in quantitative reasoning tasks.} These limitations highlight the need for a new system capable of effectively processing raw sensor inputs at scale and fusing them with practical user queries.}
To address this, we propose \Method, a novel system with a three-stage pipeline that includes a dedicated sensor data query stage. As shown in Fig.~\ref{fig:variants}, \Method significantly improves answer accuracy across all categories, particularly for quantitative tasks such as day comparisons and time queries.
\section{\Method Design}
\label{sec:model}

\subsection{Overview of \Method}

%
We design \Method, a novel end-to-end QA system for answering both quantitative and qualitative questions using long-duration, high-frequency raw sensor data. \Method achieves the goal by having a three-stage pipeline: \textit{question decomposition, sensor data query}, and \textit{answer assembly}.
\revise{\Method addresses all three challenges discussed in Sec.~\ref{sec:intro}, i.e., processing long-duration, high-frequency sensor data, effectively fusing sensor data with text questions, and generating accurate, semantically meaningful answers.
The sensor query stage is the \textit{key novel component} of \Method.
In the query stage, \Method pretrains a sensor encoder and a text encoder offline to effectively associate high-frequency and long-duration multimodal sensor timeseries with semantically meaningful embeddings. To effectively fuse sensor and text modalities, \Method further designs a custom online query mechanism using a similarity search, retrieving all sensor information relevant to the original question. This enables precise extraction of sensor context across varying durations and question types, significantly improving the accuracy of the final answer.
Last but not least, \Method utilizes two LLMs: one in the question decomposition stage to interpret user queries, and the other in the answer assembly stage to generate the final response, further enhanced through in-context learning and fine-tuning techniques.}


\begin{figure*}[tb]
  \centering
  \includegraphics[width=0.9\textwidth]{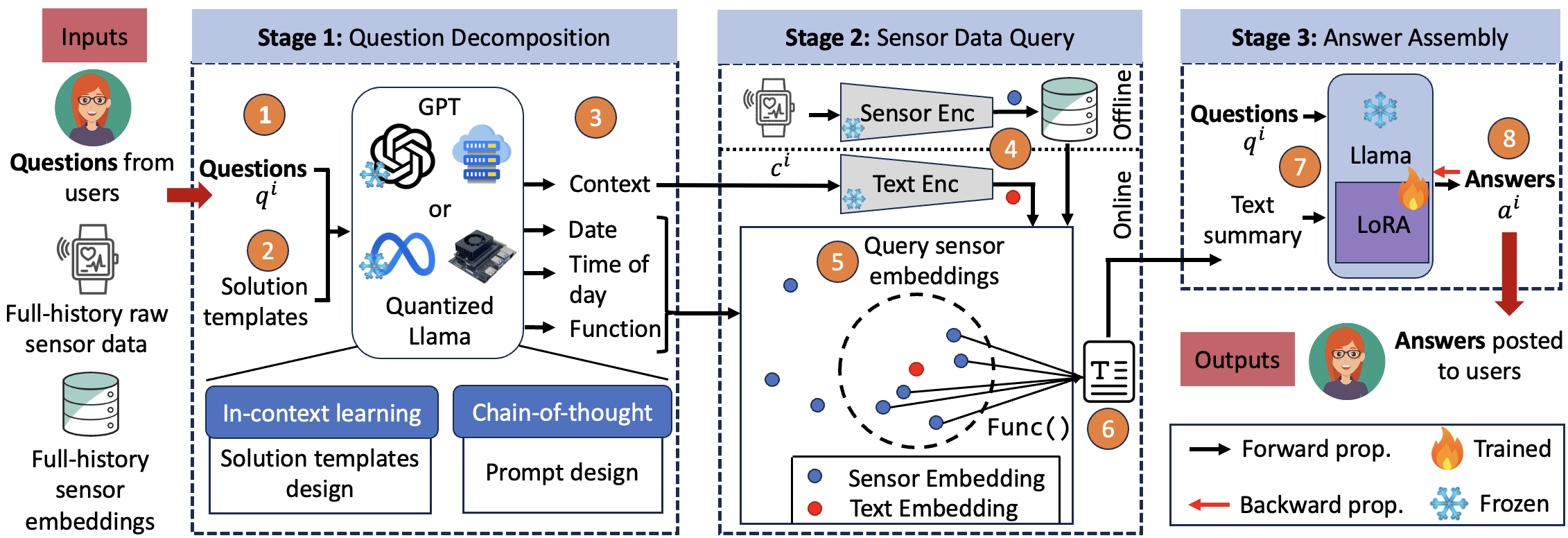}
  \vspace{-3mm}
  \caption{The system diagram of \Method including three stages. \revise{When a user asks a question, \Method first decomposes the question (stage 1), queries the relevant sensor embeddings (stage 2), and generates accurate, semantically meaningful answers (stage 3). The sensor and text encoder used in the query stage are pretrained offline in advance.}}
  \vspace{-5mm}
  \label{fig:overview}
\end{figure*}

The complete pipeline of \Method is illustrated in Fig.~\ref{fig:overview}. For example, given a question like, “\textit{How long did I exercise last week in the morning?}” (\textcircled{1} in Fig.~\ref{fig:overview}), \Method first decomposes the question and generates specific sensor data queries using LLMs. These queries include details such as the context of interest, date and time ranges, and the \revise{query} function (\textcircled{3}). In this case, the query might specify a context of “\textit{do exercise},” a time span of “\textit{last week},” a time of day of “\textit{morning},” and a \revise{query} function of \texttt{CalculateDuration}. To improve the accuracy of question decomposition, \Method uses solution templates \revise{for in-context learning} (\textcircled{2}) with carefully designed prompts. \revise{In-context learning is chosen because it enhances the quality of LLM outputs using only a small number of solution templates, compared to finetuning~\cite{fu2023gpt4aigchip}.} Next, the sensor query stage encodes the \revise{context} label “\textit{exercise}” into a text embedding (\textcircled{4}) and retrieves sensor embeddings that are sufficiently similar to the text embedding (\textcircled{5}). These sensor embeddings are encoded offline from the full-history raw sensor data. Both the sensor and text encoders are pretrained offline to align their outputs in the same embedding space, ensuring accurate sensor data retrieval.
Therefore, only text encoding and similarity-based query searches are performed online.
Additional properties, such as the date (“\textit{last week}”) and the time of day (“\textit{morning}”), are used to constrain the query range. The final step of the sensor data query stage involves summarizing the relevant sensor embeddings into text \revise{using the corresponding query function} (\textcircled{6}). For instance, with the \texttt{CalculateDuration} function, this step calculates the total duration of the retrieved sensor embeddings. This results in a sensor context such as: “\textit{Among all days last week, you exercised for 35 minutes on Monday morning and 55 minutes on Thursday morning.}” Finally, the original question and the sensor context are sent to the answer assembly stage, where a fine-tuned LLM generates the final answer (\textcircled{7}). \revise{Finetuning is used to align the model's output with human-generated answers from state-of-the-art datasets.} With accurate sensor context, the model generates a precise response, such as: “\textit{You exercised a total of 1 hour and 30 minutes last week.}” (\textcircled{8}).
\revise{In the following lines, we describe the detailed designs in each stage of \Method: question decomposition (Sec.~\ref{sec:question-decomposition}), sensor data query (Sec.~\ref{sec:sensor-data-query}) and answer assembly (Sec.~\ref{sec:answer-assembly})).}

\vspace{-3mm}
\subsection{Question Decomposition}
\label{sec:question-decomposition}

The question decomposition stage processes the user question and identifies specific query triggers for the sensor data query stage, such as the context of interest, date and time ranges, and the \revise{query} function, as illustrated in the leftmost box in Fig.~\ref{fig:overview}.
We choose to rely on LLMs for this stage due to their outstanding capability in handling natural language inputs.
However, LLMs are not without limitations as they can produce erroneous outputs or hallucinations~\cite{huang2023survey}.
The key challenge in designing this stage lies in generating \textit{accurate} decompositions for \textit{arbitrary} user inputs, \revise{particularly in the absence of a dedicated dataset for this task}. \revise{To address this, \Method leverages the capabilities of a pre-trained LLM by employing in-context learning~\cite{shao2023prompting} and chain-of-thought techniques~\cite{chuCoTReasoningSurvey2024}. These approaches are chosen because, compared to finetuning~\cite{hu2021lora}, they enhance the quality of LLM outputs using a small number of manually crafted examples, eliminating the need for a high-quality, task-specific dataset.}

\begin{figure}[t]
    \centering    
    \begin{subfigure}[b]{0.9\textwidth}
        \centering
        \includegraphics[width=\textwidth]{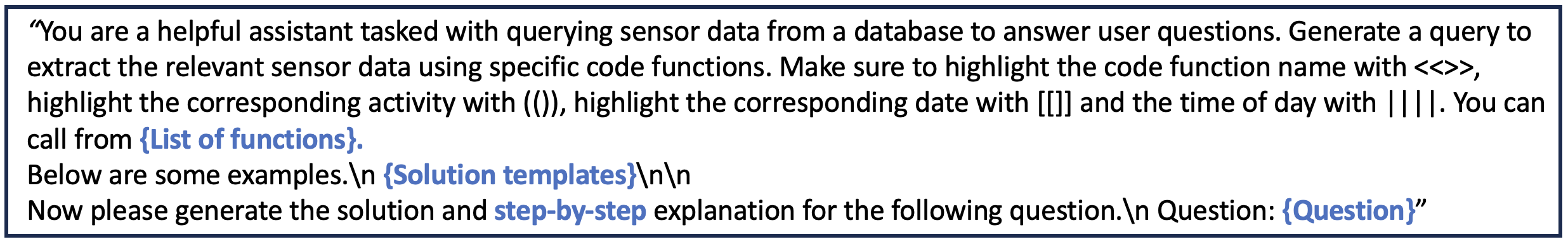}
        \vspace{-6mm}
        \caption{The prompt design for the question decomposition stage includes ICL and CoT. \revise{A list of available query functions, solution templates, and the user’s question need to be included in the prompt.}}
        \label{fig:question-decompose}
    \end{subfigure}

    \begin{subfigure}[b]{0.9\textwidth}
        \centering
        \includegraphics[width=\textwidth]{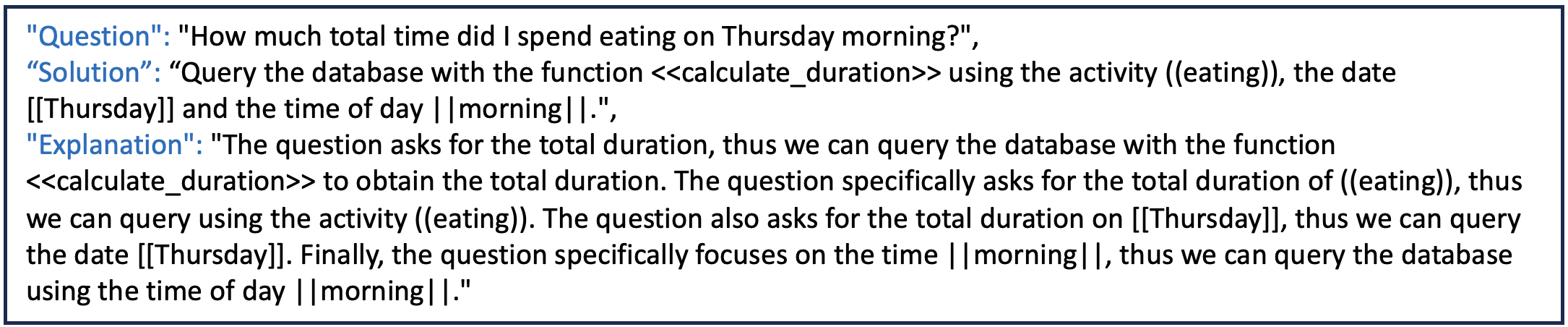}
        \vspace{-6mm}
        \caption{Example solution template for a time-related query to be inserted into the prompt to question decomposition.}
        \label{fig:solution-template}
    \end{subfigure} 

    \vspace{-4mm}
    \caption{Prompts design in the question decomposition stage of \Method to enhance the generating quality of LLMs.}
    \label{fig:prompts}
    \vspace{-5mm}
\end{figure}

An example prompt for question decomposition is illustrated in Fig.~\ref{fig:question-decompose}. We instruct the LLM to mark different arguments with distinct symbols, such as, using "<<" and ">>" for function names, "((" and "))" for context labels, etc. This enables accurate extraction of multiple arguments from a single LLM interaction. 
\Method also supports multiple extracted phrases to handle complex questions such as ``\textit{How long did I work at school on Monday and Tuesday?}''
The question decomposition stage is designed with the flexibility to support new extractable properties.
We next explain more details on the in-context learning and chain-of-thought techniques. 

\textbf{In-Context Learning (ICL)} integrates a few examples directly into the prompt during inference, enabling LLMs to adapt effectively to specific tasks without requiring fine-tuning~\cite{alayrac2022flamingo,shao2023prompting}. 
Building on this insight, we design a library of general solution templates covering diverse QA scenarios and incorporate them as in-context examples to improve decomposition accuracy (Fig.~\ref{fig:question-decompose}).
We observe that decomposition solutions for the same question type often share similarities. For instance, "how long" questions typically map to the \texttt{CalculateDuration} function. To utilize this question-specific property, in \Method, we first classify the question type using a BERT model~\cite{devlin2018bert}. Then, we dynamically select templates that best match the question category. \revise{This strategy has been shown to improve ICL performance, even in settings that are different from ours but face the same challenge of limited annotated data~\cite{fu2023gpt4aigchip}.}
Fig.~\ref{fig:solution-template} illustrates an example of a solution template for a time query. 

\textbf{Chain-of-Thought (CoT) Prompting} explicitly requests LLMs to generate their reasoning process, which improves accuracy in zero- or few-shot logical reasoning~\cite{chuCoTReasoningSurvey2024,lu2022learn}. In \Method, we include ``\textit{please generate step-by-step explanations}'' in prompts to encourage CoT, as shown in Fig.~\ref{fig:question-decompose}.
CoT and detailed solution templates enhance logical reasoning, especially for multi-step questions.

\subsection{Sensor Data Query}
\label{sec:sensor-data-query}

\revise{The sensor query stage, as the most important module in \Method, is responsible for processing the long-duration, high-frequency raw sensor signals and extracting the most relevant information given arbitrary user queries. This enables effective fusion of sensor data with natural language. As shown in the middle part of Fig.~\ref{fig:overview}, the sensor data query stage takes the extracted arguments from question decomposition, such as the context of interest (e.g., “\textit{do exercise}”), time span (e.g., “\textit{last week}”), time of day (e.g., “morning”) and query function (e.g., \texttt{CalculateDuration}). The output of the stage is a concise summary of the relevant sensor data in text form.
\Method introduces two novel components to addresses these challenges. First, during offline training, \Method pretrains a sensor encoder and a text encoder using a novel contrastive loss to align sensor signals with partial context (Sec.~\ref{sec:pretraining}).  Then, during online querying, \Method designs a set of custom query functions to extract relevant sensor embeddings, which are converted to textual summarizations for the final answer assembly stage (Sec.~\ref{sec:online-query}).}

\vspace{-2mm}
\subsubsection{Contrastive Sensor-Text Pretraining for Partial Contexts}
\label{sec:pretraining}
The sensor and text encoders are crucial for effective sensor-text fusion, serving as a "bridge" between high-frequency sensor timeseries and semantic text. 
\revise{A key challenge in \Method is ensuring that the sensor and text encoders can handle diverse context phrases extracted from user questions, which may vary unpredictably based on users' habits. Suppose $c_i$ represents the context argument extracted from question $q_i$. Since \Method is designed for general daily life monitoring, $c_i$ can refer to anything from activities, locations, or social interactions (e.g., spending time with family)., depending on the user’s interest. For example, in the question “\textit{Did I go to school?}”, $c_i$ is “\textit{go to school,}” which refers to a location. This means the encoders in \Method must be able to align sensor data with a broad range of contexts.}
However, achieving this level of alignment is difficult. Existing pretraining methods like CLIP and its variants~\cite{radford2021learning,moon-etal-2023-imu2clip} work well when aligning sensor data with full, detailed sentences (e.g., “\textit{The person is sitting and working on computers at school.}”). However, when a user is specifically interested in a partial context like "\textit{working on computers,}" the encoded text embedding may not closely match the original sensor embeddings, leading to reduced accuracy in sensor data query as we show in Appendix~\ref{sec:ablation}. To address this challenge, we introduce a novel contrastive sensor-text pretraining loss for partial contexts.

\begin{figure}[t]
\begin{center}
\includegraphics[width=0.75\textwidth]{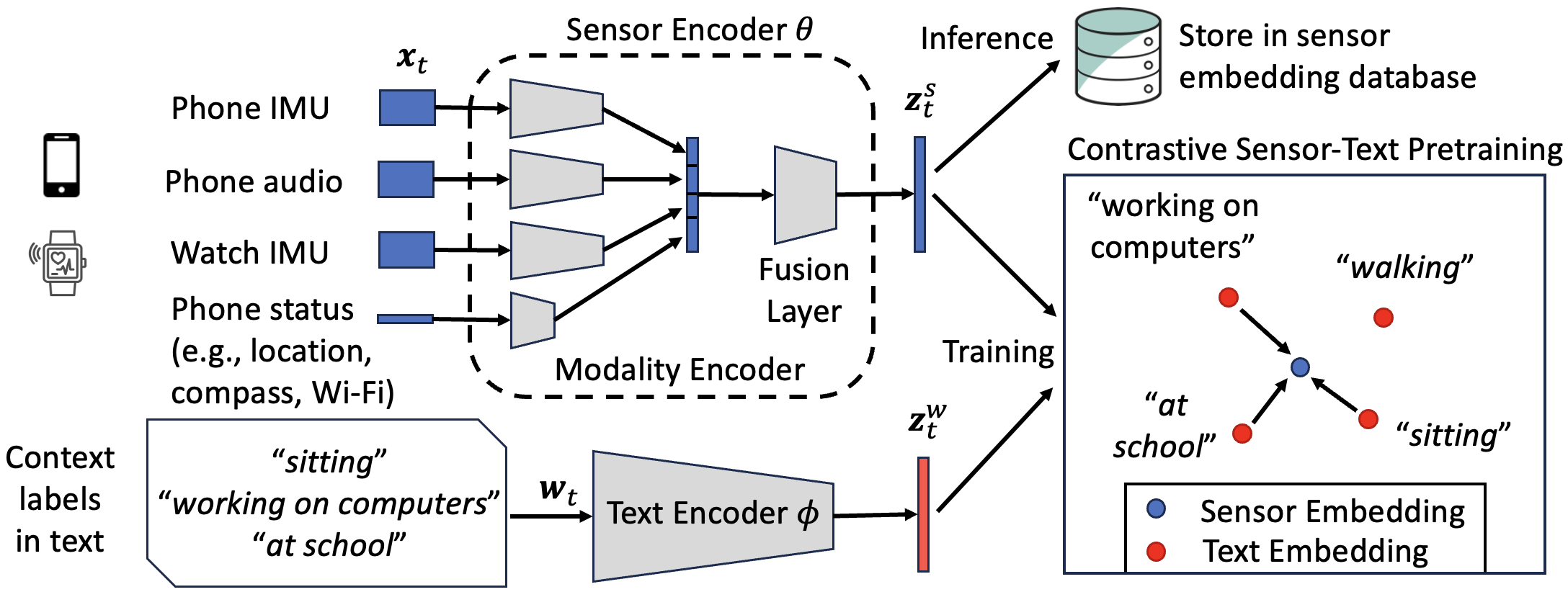} 
\vspace{-4mm}
\caption{Visualization of the contrastive sensor-text pretraining in \Method on a sample $\{\mathbf{x}_t, w_t\}$. \revise{At time $t$, the raw sensor input $\mathbf{x}_t$ includes data from phone and watch IMUs, audio, and phone status. The time series signals are split into time windows. The positive context labels are $w_t = \{ \textrm{sitting, working on computers, at school}\}$ while other labels such as “walking” are negative ones. Our pretraining loss increases the similarity between $\mathbf{x}_t$ and the positive labels, and reduces it for the negative ones.}}
\label{fig:sensor_encoder}
\end{center}
\vspace{-6mm}
\end{figure}

\revise{Fig.~\ref{fig:sensor_encoder} presents the details of our pretraining process.} We pretrain our model on a large-scale multimodal sensor dataset with annotations, where each sample $\{\mathbf{x}_t, w_t\}$ consists of one segment of raw time series sensor data $\mathbf{x}_t$ collected at time $t$ and an associated set of partial context labels $w_t$. $w_t$ can be extracted from label annotations. For instance, in a single-label human activity classification dataset,  $w_t$ may contain a single phrase (e.g., $\{\textrm{"walking"}\}$), whereas in a multi-label dataset, it may include multiple phrases (e.g., $\{ \textrm{"at school", "working on computers"}\}$).
We employ separate encoders for sensor and text inputs. 
Formally, let $\theta$ denote the complete sensor encoder model. The encoded sensor embedding is given by $\mathbf{z}^s_t = \theta(\mathbf{x}_t)$.
The text encoder, denoted by $\phi$, maps arbitrary phrases to a text embedding $\mathbf{z}^w_t = \phi(w_t)$.
Key notations used throughout our work are summarized in Table~\ref{tbl:notation}.
We use distinct sensor encoder for each sensor modality (e.g., IMUs, audio and phone status) to accommodate the varying complexity of different sensor data types. For instance, a Transformer-based encoder is used for high-frequency time series data while a simple linear layer is designed for encoding phone status. The outputs from these modality-specific encoders are concatenated and passed through a fusion layer to generate the final sensor embedding. Our framework allows for missing modalities by padding with mean values and is flexible for future expansion to additional sensor modalities.

%
We introduce a new pretraining loss to align sensor embeddings $\mathbf{z}^s_t$ and partial text embeddings $\mathbf{z}^w_t$. Different from CLIP~\cite{radford2021learning}, our loss function enables alignment between sensor data and all partial phrases:
\begin{equation}
 \mathcal{L} = \sum_t {\frac{-1}{|w_t|} \sum_{w \in w_t} \log \frac{\exp(\mathbf{z}^s_t \cdot \mathbf{z}^w_t / \tau)}{\sum_{a \in A(w)} \exp (\mathbf{z}^s_t \cdot \mathbf{z}^a / \tau)}}, \label{eq:loss} 
\end{equation}
$|w_t|$ is the cardinality of set $w_t$, $\tau$ is a scalar temperature parameter. The term $A(w) \equiv A \setminus \{w\}$ represents the collection of negative contexts, defined as all possible phrases except the positive phrase $w$.
The intuition behind our loss function can be elaborated with a concrete example, as shown in Fig.~\ref{fig:sensor_encoder}. Consider a sensor embedding $\mathbf{x}_t$ and a set of partial text labels $w_t = \{ \textrm{sitting, working on computers, at school}\}$.
Our loss function encourages high similarity between the sensor embedding $\mathbf{z}^s_t$ and all positive text embeddings corresponding to $w_t$, namely "sitting", "working on computers", \revise{"at school"}, while distinguishing them from other negative contexts such as \revise{"walking"}. We draw inspiration from the supervised contrast learning loss~\cite{khosla2020supervised}. However, our loss function differs in that it explicitly models the similarity between sensor embeddings and multiple text phrases rather than between samples of the same modality.
After pretraining, we store all sensor embeddings in a database for online queries.

\begin{table}[t]
\footnotesize
\caption{List of important notations.}
\label{tbl:notation}
\vspace{-4mm}
\begin{center}
\begin{tabular}{p{3em} p{21em} | p{3em} p{21em} } 
\toprule
Symbol & Meaning & Symbol & Meaning \\
\midrule
 \revise{$q^i, a^i$} & \revise{Question \& answer pair} &
 \revise{$c^i$} & \revise{Extracted context from question decomposition} \\
 $\mathbf{x}_t$ & Multimodal sensor data sampled at $t$ &
 $w_t$ & Partial context labels at $t$ \\
 $d$ & Number of different sensors &
 $T$ & Total duration of sensor data \\
 $\theta$ & Sensor encoder &
 $\phi$ & Text encoder \\
 $\mathbf{z}^s_t$ & Sensor embedding of $\mathbf{x}_t$ &
 $\mathbf{z}^w_t$ & Text embedding of $w_t$ \\
 $A(w)$ & Set of negative contexts of $w$ &
 $\tau$ & Temperature scalar \\
 $f$ & Trained similarity function between embeddings & 
 $h$ & Threshold in query search \\
\bottomrule
\end{tabular}
\end{center}
\vspace{-4mm}
\end{table}

\vspace{-2mm}
\subsubsection{Online Sensor Data Query}
\label{sec:online-query}
\Method further introduces a query mechanism to extract relevant sensor embeddings and convert them into textual summarizations based on the pretrained sensor and text encoders.
\revise{The active query process of \Method differs significantly from the passive journaling or summarization modules~\cite{xu2024autolife,ouyang2024llmsense}, as \Method dynamically selects the query functions and query ranges based on the questions.
The performance of sensor data query directly impacts the accuracy of \Method's answers.}
As shown in the middle part of Fig.~\ref{fig:overview}, after \Method encodes the context $c_i$ using the text encoder (\textcircled{4}), \Method performs a similarity search among the sensor embedding given the text embedding (\textcircled{5}), and summarizes the retrieved sensor embeddings into textual information \revise{using custom query functions} (\textcircled{6}).
We next detail the similarity-based embedding search and query function design.

\textbf{Similarity-based Embedding Search} 
Our goal is to accurately identify all sensor samples relevant to the query context.
For example, if the query context $c_i$ is ``\textit{exercise}'', \Method aims to retrieve all sensor samples associated with activities similar to exercise.
If multiple text are generated during question decomposition, \Method performs an embedding search for each query text individually.
The pretraining in Sec.~\ref{sec:pretraining} ensures that the sensor embedding space is well aligned with the text embedding space even using partial context. 
Therefore, the similarity comparison given any pair of sensor and text embeddings can be achieved by training a similarity function $f$ in the embedding space: 
\begin{equation}
f(\mathbf{z}^s_t, \mathbf{z}^w) = f \big(\theta(\mathbf{x}_t), \phi(w) \big) \in [0, 1]
\end{equation}
The output of $f$ is a scalar value between 0 and 1, representing the similarity between the sensor and text samples.
With $f$, the similarity search in the embedding space is significantly more efficient than searching directly through raw, high-frequency time-series data. The efficiency can be further improved by narrowing the search scope based on the date and time-of-day arguments identified during question decomposition.

\textbf{\revise{Query} Function Design}
\Method uses a set of \revise{query} functions to filter the relevant sensor samples and generate textual information for the answer assembly stage. The specific \revise{query} function to be used is determined during question decomposition.
For example, a question of "\textit{how long}" should be directed to the \texttt{CalculateDuration} function, while a question of "\textit{what did I do}" should be handled by the \texttt{DetectActivity} function.
Each \revise{query} function uses a unique template to return text information.
The numerical value in the returned text is determined by the embedding search results.
For instance, when querying \texttt{CalculateDuration} with the activity ``\textit{cooking}'', date ``\textit{Sunday}'', and time ``\textit{morning}'', the textual output can be: "\textit{You spent {$\gamma$} minutes cooking on Sunday morning}", where $\gamma$ is calculated as follows:
\begin{equation}
\gamma = \sum_{t \in T_{SundayMorning}} \Bigg[ f \big( \theta(\mathbf{x}_t), \phi( \textrm{"cooking"}) \big) > h \Bigg].
\end{equation}
Here $T_{SundayMorning}$ represents all timestamps within Sunday morning. 
The notation $[Cond]$ gives $1$, when the inner condition {\it Cond} is met; otherwise $0$.
$h$ is a predetermined threshold.
In \Method, we carefully design a set of \revise{query} functions to account for diverse scenarios in real life, including time quries, activity quries, counting, etc. The details of those functions are explained in Appendix~\ref{sec:query-function}.

\vspace{-2mm}
\subsection{Answer Assembly}
\label{sec:answer-assembly}

As shown in the right box of Fig.~\ref{fig:overview}, the final stage of answer assembly integrates question and sensor information to generate the final answer. 
\revise{State-of-the-art methods~\cite{xing2021deepsqa,zhang2023llama,moon-etal-2023-imu2clip} rely on training a neural network to fuse natural language and sensor data, similar to a ``black-box''. These methods often lead to ineffective fusion and inaccurate answers when dealing with long-duration, high-frequency sensor data (see Sec.~\ref{sec:motivation}).}
In contrast, \Method directly fuses the filtered senor context from the query stage with the original question.
Our intuition is that, in contrast to processing and fusing with other modalities, LLMs are the most professional in dealing with text.
\Method is capable of answering both qualitative and quantitative questions by combining the original question and the extracted fine-grained activity information from long-duration, high-frequency sensors.
\revise{At this answer assembly stage, we finetune a LLM such as LLaMA~\cite{zhang2023llama} to adapt the model to the desired answer style. Fine-tuning is chosen over few-shot learning as it delivers better performance with the presence of high-quality datasets like \Dataset~\citesensorqa, which is verified in Appendix~\ref{sec:ablation}.} We use Low Rank Adaptation (LoRA)~\cite{hu2021lora} due to its parameter efficiency and comparable performance to full fine-tuning.
\revise{The prompt is as follows: \textit{``Answer the question based on the context below.
Context: \{Context\} Question: \{Question\} Response:''}}

\vspace{-2mm}
\section{System Implementation}
\label{sec:system-implementation}

\begin{figure}[t]
  \begin{subfigure}[b]{0.38\textwidth}
        \centering
        \includegraphics[width=\textwidth]{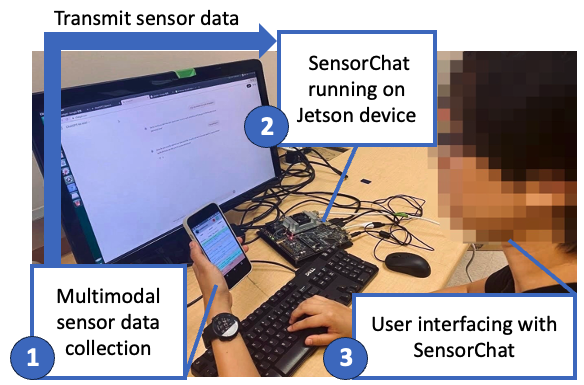}
        \vspace{-5mm}
        \caption{Real system implementation.}
        \label{fig:system_diagram}
    \end{subfigure} 
    \begin{subfigure}[b]{0.3\textwidth}
        \centering
        \includegraphics[width=\textwidth]{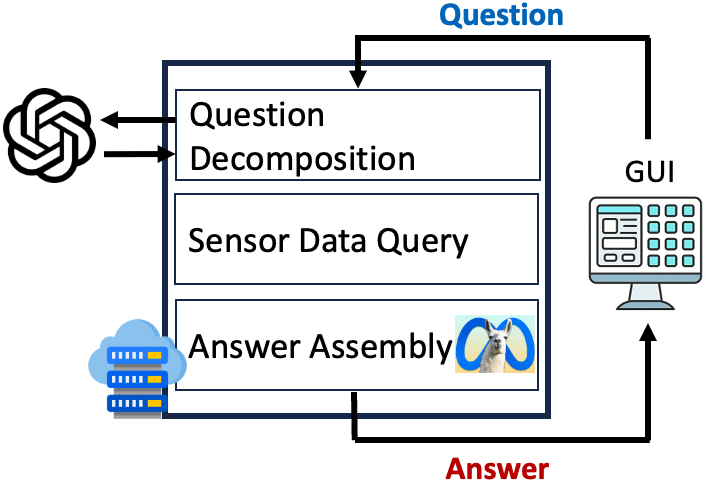}
        \vspace{-5mm}
        \caption{\MethodC diagram.}
        \label{fig:sensorchat_cloud}
    \end{subfigure}
    \begin{subfigure}[b]{0.28\textwidth}
        \centering
        \includegraphics[width=\textwidth]{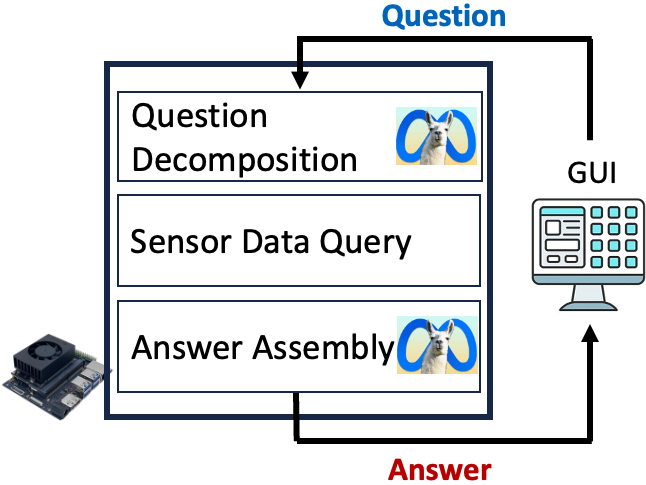}
        \vspace{-5mm}
        \caption{\MethodE diagram.}
        \label{fig:sensorchat_edge}
    \end{subfigure}
    \vspace{-4mm}
    \caption{System implementation of \Method.}
    \label{fig:system}
    \vspace{-6mm}
\end{figure}

We implement \Method on real-world systems. We envision \Method as a personal assistant that provides accurate and timely answers to user questions, as outlined in our problem statement (Sec.~\ref{sec:problem-statement}).
Fig.~\ref{fig:system} visualizes the general system pipeline of \Method.
We employ smartphones and smartwatches to collect multimodal sensor data from users in daily lives (\textcircled{1} in Fig.~\ref{fig:system_diagram}).
Implemented based on the ExtraSensory App~\cite{vaizman2018extrasensory}, the mobile devices automatically gathers data for 20 seconds every minute, including 40Hz IMU signals, 13 MFCC audio features from a 22kHz microphone, and other phone state information including compass, GPS location, Wi-Fi status, light intensity, battery level, etc. The details can be found in the ExtraSensory App manual~\cite{vaizman2018extrasensory}.
These sensor data are then transmitted to a system running \Method (\textcircled{2} in Fig.~\ref{fig:system_diagram}). We offer two variants of \Method, designed for a cloud server and an edge environment respectively. Their detailed implementations and trade-offs are discussed below. Finally, users can interact with \Method directly through a chatting interface using natural language, shown as \textcircled{3} in Fig.~\ref{fig:system_diagram}.

\textbf{\MethodC and \MethodE} We offer two system variants of \Method, as shown in Fig.~\ref{fig:sensorchat_cloud} and \ref{fig:sensorchat_edge}.
\begin{itemize}
    \item \textbf{\MethodC}, designed for a cloud environment, uses GPT-3.5-Turbo~\cite{gpt-3.5} for question decomposition and a full-size finetuned LLaMA2-7B model~\cite{touvron2023llama} for answer assembly. We deploy and test \Method on a cloud server equipped with an NVIDIA A100 GPU~\cite{a100}.
    \item \textbf{\MethodE}, designed for an edge environment, uses quantized LLaMA model~\cite{touvron2023llama} for both question decomposition and answer assembly. The question decomposition model is quantized from the official LLaMA3-8B, while the answer assembly model is quantized from our fine-tuned version of LLaMA2-7B. 
    We use Activation-aware Weight Quantization (AWQ), a state-of-the-art quantization method for LLMs, known for its hardware efficiency. We deploy and test \MethodE on a NVIDIA Jetson Orin NX module~\cite{jetsonorin} with 16GB RAM.
\end{itemize}
\MethodC and \MethodE accommodate two typical use scenarios. \MethodC is expected to deliver superior QA performance with the full-precision LLMs in the cloud, at the cost of intensive resource consumption. Additionally, \MethodC requires the users to transmit the full sensor history to the cloud server.
On the other hand, \MethodE runs entirely on a local edge platform belonging to the user, eliminating the need to transmit user data to the cloud and thus preserving user privacy. However, its QA and latency performance degrade compared to \MethodC.  
\revise{Further implementation details can be found in Appendix~\ref{appendix:implementation-details}.}


\vspace{-2mm}
\section{Evaluation on State-of-the-Art Dataset}
\label{sec:evaluation}

In this section, we thoroughly evaluate \Method on the state-of-the-art dataset focusing on quantitative questions. \revise{Due to space limitations, we defer the ablation study, sensitivity analysis, and generalization study to Appendix~\ref{sec:ablation}–\ref{sec:generalizability}.} We further validate \Method in a real world study with \revise{both quantitative and qualitative questions} in Sec.~\ref{sec:deployment}.

\vspace{-2mm}
\subsection{Dataset and Metrics}

In this evaluation section, we focus mainly on the \Dataset dataset~\citesensorqa and quantitative questions. 
To the best of our knowledge, \Dataset is the first and only available benchmarking dataset for QA interactions that use long-term timeseries sensor data and reflect practical user interests.
While we focus on \Dataset~\citesensorqa in this section, we emphasize that \Method is broadly applicable and can be extended to other real-world sensing applications.
To ensure the best alignment between the QA pairs and sensor information, we conduct offline encoders pretraining on the ExtraSensory multimodal sensor dataset~\cite{vaizman2017recognizing}, which servers as the sensor data source for \Dataset.
During pretraining, all sensor samples are aligned by a time window of 20 seconds.


\textbf{Dataset Variants and Metrics}
We evaluate three versions of \Dataset~\cite{sensorqa} using various metrics to assess both the quality and accuracy of the generated answers.
\begin{itemize}[topsep=0pt, itemsep=0pt]
    \item \textbf{Full answers} refer to the original full responses  in \Dataset. We evaluate the model's performance on the full answers dataset using Rouge-1, Rouge-2, and Rouge-L scores~\cite{eyal-etal-2019-question}. Rouge scores measure the overlap of n-grams between the machine-generated content and the ground-truth answers, expressed as F-1 scores. Higher Rouge scores indicate greater similarity between the generated and true answers.
    \item \textbf{Short answers} are the 1-2 key words extracted from the full answers by GPT-3.5-Turbo~\cite{gpt-3.5}, offered with the original \Dataset dataset~\citesensorqa. We use the exact match accuracy on the short answers to evaluate the precision of generated answers, as detailed in Sec.~\ref{sec:motivation}.
    \item \textbf{Multiple choices} are generated by prompting GPT-3.5-Turbo~\cite{gpt-3.5} to create three additional choices similar to the correct short answer. An example QA can be "Which day did I spend the most time with coworkers? A. Friday, B. Monday, C. Thursday, D. Wednesday", with the correct answer being "D" or "D. Wednesday." The models are expected to accurately select the correct answer from the four candidates. We evaluate the performance based on exact answer selection accuracy.
\end{itemize}
We create the multiple-choice version in addition to the full answers and short answers provided in the original \Dataset dataset~\citesensorqa, to further assess the model's ability in distinguishing similar facts based on sensor data. 

\subsection{State-of-the-Art Baselines}

We compare \Method against the state-of-the-art baselines that leverage different modalities combinations, including \textcolor{mygreen}{text}, \textcolor{myred}{vision+text}, and \textcolor{myblue}{sensor+text} data. 
\revise{For the text-only models, we employ the popular \textbf{GPT-4~\cite{gpt-4}}, \textbf{T5~\cite{2020t5}} and \textbf{LLaMA~\cite{touvron2023llama}} to obtain a baseline performance without considering the sensor data. 
For the vision+text models, we adopt \textbf{GPT-4-Turbo~\cite{gpt-4}, GPT-4o~\cite{gpt-4}}, \textbf{LLaMA-Adapter~\cite{zhang2023llama}} and \textbf{LLaVA-1.5~\cite{liu2024improved}}. We feed the activity graphs in \Dataset~\citesensorqa (similar to Fig.~\ref{fig:sensorchat_example}) into these models along with the questions as input. We also compare against the latest sensor+text models including \textbf{IMU2CLIP+GPT-4o~\cite{moon-etal-2023-imu2clip}}, \textbf{LLMSense~\cite{ouyang2024llmsense}}, \textbf{DeepSQA~\cite{xing2021deepsqa}} and \textbf{OneLLM~\cite{han2023onellm}}. To ensure a fair comparison, we adapt these baselines to process the same long-duration, high-frequency data by modifying the data format to match their original input modality. Further adaptation details are provided in Appendix~\ref{appendix:baseline-implementation-details}.}

For the closed-source generative baselines based on GPT, we use few-shot learning (FSL). Specifically, we incorporate a set of QA examples from \Dataset~\citesensorqa into the prompt for each question input. We adopt $10$ samples per question based on a grid search of $\{2, 5, 10, 15\}$.
For T5 and DeepSQA, we train the models directly on the \Dataset dataset~\citesensorqa.
For open-source LLM baselines, we apply LoRA fine-tuning (FT)~\cite{hu2021lora} using the samples from \Dataset~\citesensorqa. 
We randomly select 80\% of the QA pairs in \Dataset~\citesensorqa as training samples and reserve the remaining 20\% for testing. We explore alternative splitting schemes in Appendix~\ref{sec:generalizability} to demonstrate \Method's generalizability to unseen users.
All baselines based on LLaMA (that is, LLaMA, LLaMA-Adapter, and OneLLM) use LLaMA2-7B~\cite{touvron2023llama}.
All baselines adopt the same hyperparameters as those specified in their official codebases.
%

\begin{figure*}[tb]
  \centering
  \centering
  \includegraphics[width=0.82\textwidth]{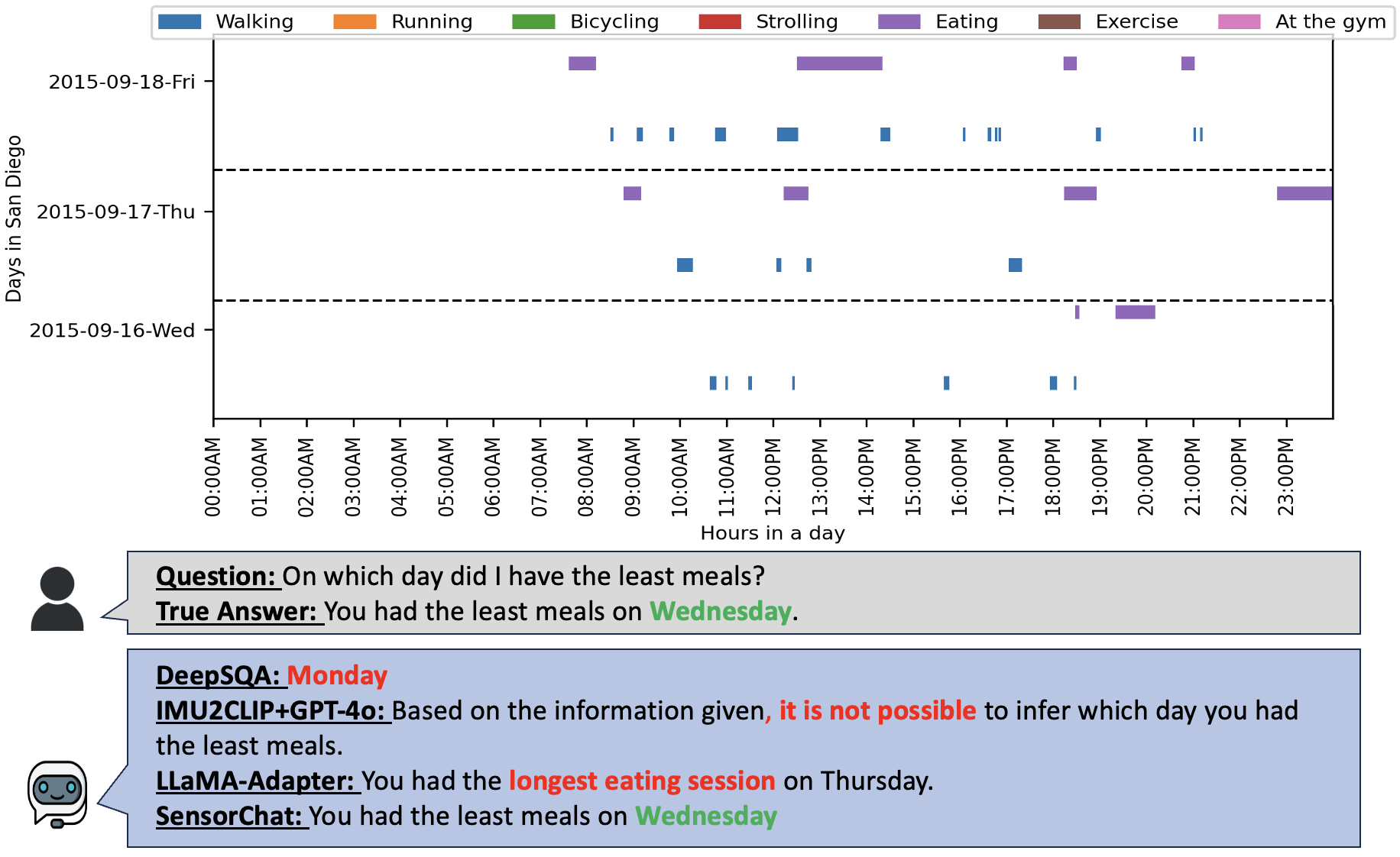}
  \vspace{-4mm}
  \caption{Qualitative results of \Method in comparison to state-of-the-art methods on \Dataset~\citesensorqa. \revise{Key phrases in the answers are highlighted in the green if they match the true answer, and in the red if they do not. \Method is the only method among the presented baselines that correctly interprets the question and generates an accurate answer.}}
  \vspace{-4mm}
  \label{fig:qual_results}
\end{figure*}

\subsection{QA Performance}
\label{sec:qa-performance}

\revise{We start with presenting an example of \Method compared to top-performing baselines in Fig.~\ref{fig:qual_results}. Additional QA examples are presented in Appendix~\ref{appendix:qualitative}.}
\revise{Answering the question in Fig.~\ref{fig:qual_results} involves two steps: counting meal frequency per day and identifying the day with the fewest meals. All three baselines fail in this challenging task. DeepSQA~\cite{xing2021deepsqa} and LLaMA-Adapter~\cite{zhang2023llama} struggle with long-duration, high-frequency data, as they were originally designed for short sensor inputs under one minute. In our setting, these models fail to learn effective fusion between long-term sensor data and textual questions, leading to wrong answers.
IMU2CLIP+GPT-4o~\cite{moon-etal-2023-imu2clip} works by first mapping raw sensor signals into a long narrative log and then passing the log to GPT-4o for reasoning.
Although IMU2CLIP+GPT-4o is able to accurately associate sensor signals with text, it is constrained by GPT-4o’s token limit, causing key portions of the log to be truncated and leading to incomplete or uninformative answers.}
In contrast, \Method succeeds by decomposing the query in the first stage, extracting daily meal frequencies in the second stage, and leaving the comparison to the answer stage.
The collaboration across the three stages allows \Method to effectively handle quantitative queries that require precise numerical reasoning across multi-day span, which highlights \Method's advancements over existing works.

\begin{table*}[t]
{
\footnotesize
\centering
\begin{tabular}{c|c|c|ccc|c|c}
\toprule
 \textbf{Modalities} & \textbf{Method} & \textbf{FSL/FT$^1$} & \multicolumn{3}{c|}{\textbf{Full Answers}} & \textbf{Short Answers} & \textbf{Multiple Choices} \\
 & & & Rouge-1 ($\uparrow$) & Rouge-2 ($\uparrow$) & Rouge-L ($\uparrow$) & Accuracy ($\uparrow$) & Accuracy ($\uparrow$) \\
\midrule
\textcolor{mygreen}{Text} & GPT-4~\cite{gpt-4} & FSL & 0.66 & 0.51 & 0.64 & 0.16 & 0.34 \\
\textcolor{mygreen}{Text} & T5-Base~\cite{2020t5} & FT & 0.71 & 0.55 & 0.69 & 0.25 & 0.52 \\
\textcolor{mygreen}{Text} &  LLaMA2-7B~\cite{llama2} & FT & \underline{0.72} & \underline{0.62} & \underline{0.72} & 0.26 & \underline{0.56} \\
\hline
\textcolor{myred}{Vision+Text} & GPT-4-Turbo~\cite{gpt-4} & FSL & 0.38 & 0.28 & 0.36 & 0.14 & 0.24 \\
\textcolor{myred}{Vision+Text} & GPT-4o~\cite{gpt-4} & FSL & 0.39 & 0.28 & 0.37 & 0.20 & 0.07 \\
\textcolor{myred}{Vision+Text} & LLaMA-Adapter~\cite{zhang2023llama} & FT & 0.73 & $0.57$ & $0.71$ & \underline{0.28} & 0.54 \\
\textcolor{myred}{Vision+Text} & LLaVA-1.5~\cite{liu2024improved} & FT & $0.62$ &  $0.46$ & $0.60$ & 0.21 & 0.47 \\
\hline
\textcolor{myblue}{Sensor+Text} & \revise{IMU2CLIP+GPT4o~\cite{moon-etal-2023-imu2clip}} & FSL & \revise{0.58} & \revise{0.44} & \revise{0.55} & \revise{\underline{0.28}} & \revise{0.49} \\ 
\textcolor{myblue}{Sensor+Text} & \revise{LLMSense~\cite{ouyang2024llmsense}} & FSL & \revise{0.37} & \revise{0.26} & \revise{0.35} & \revise{0.23} & \revise{0.49} \\ 
\textcolor{myblue}{Sensor+Text} & DeepSQA~\cite{xing2021deepsqa} & FT & 0.34 & 0.05 & 0.34 & 0.27 & - \\
\textcolor{myblue}{Sensor+Text} & OneLLM~\cite{han2024onellm} & FT & 0.12 & 0.04 & 0.12 & 0.05 & 0.30 \\
\midrule
\textcolor{myblue}{Sensor+Text} & \textbf{\MethodC} & FT & \textbf{0.77}& \textbf{0.62} & \textbf{0.75} & \textbf{0.54} & \textbf{0.70} \\
\textcolor{myblue}{Sensor+Text} & \textbf{\MethodE} & FT & 0.76 & 0.60 & 0.74 & 0.49 & 0.67 \\
\bottomrule
\multicolumn{8}{l}{$^{1}$\small{FS: Few-Shot Learning. FT: Finetuning.}} \\
\end{tabular}
}
\vspace{-1mm}
\caption{Comprehensive results of \Method compared against state-of-the-art methods on \Dataset~\citesensorqa. Bold and underlined values show the best results (all achieved by \Method) and the best among baselines. \revise{DeepSQA~\cite{xing2021deepsqa} was not evaluated on the multiple-choice version due to its inability to handle dynamic answer choices.}} 
\vspace{-7mm}
\label{tab:quant_results}
\end{table*}


Table~\ref{tab:quant_results} presents the comprehensive results of all methods on the three variants of \Dataset~\citesensorqa: full answers, short answers, and multiple-choice. 
\revise{The full answers dataset evaluates overall language quality, while the short answers and multiple-choice evaluations focus on the model's ability to learn underlying facts and numerical results rather than the language itself.}
Our \Method outperforms the best state-of-the-art methods with the \textbf{highest Rouge scores on full answers}, \textbf{\revise{93\%} higher accuracy on short answers}, and \textbf{\revise{25\%} higher accuracy on multiple choices}.
\MethodC performs slightly better than \MethodE mainly due to the stronger model capability of GPTs compared to quantized LLaMA in question decomposition. However, \MethodE, as a pure edge solution, preserves better user privacy as discussed in Sec.~\ref{sec:system-implementation}.
\revise{The gains of \Method over the baselines are more significant for short answers/multiple choices than for full answers, highlighting \Method's particular advantage in factual and numerical reasoning in comparison to existing works.}
It is important to note that the exact match accuracy for short and multiple-choice answers is a strict metric, as it requires the model to generate the \textit{exact} answer as the correct one. For instance, "4 hours" and "3 hours 50 min" would be considered different. Answering multiple-choice questions can also be particularly challenging, as candidate answers differ only slightly, such as "A. 10 min" vs. "B. 20 min," making them difficult to distinguish. In practical applications, however, a QA system does not necessarily need to achieve perfect exact match accuracy to be useful. We leave the exploration of more advanced metrics that better align with user satisfaction for future work, as discussed in Sec.~\ref{sec:future-work}.
Even under the strict exact match evaluation, \Method achieves 54\% accuracy on short answers and 70\% on multiple-choice questions. 
\textbf{These accuracy improvements demonstrate \Method's effectiveness in learning the underlying life facts from the long-duration, multimodal time series sensor data.}



\revise{In contrast, all baseline methods struggle with quantitative accuracy, achieving at most only 28\% on short-answer questions. Surprisingly, text-only baselines like LLaMA2-7B~\cite{touvron2023llama} and T5~\cite{2020t5} achieve some of the highest accuracy among the baselines, with 27\% accuracy on short answers. The limited improvements from adding vision or sensor modalities suggest that existing baselines fail to effectively fuse multimodal information in our setting.
There are several reasons for this underperformance. GPT-based models (e.g., GPT-4-Turbo and GPT-4o~\cite{gpt-4}) are not trained for interpreting sensor data or reasoning about daily activities. Their lower Rouge scores are often due to overly long answers that deviate from the ground truth. 
Vision+text models like LLaMA-Adapter~\cite{zhang2023llama} and LLaVA-1.5~\cite{liu2024improved} were originally developed for video captioning and lack the precision needed for quantitative reasoning in daily life activities.
For sensor+text baselines, IMU2CLIP+GPT-4o~\cite{moon-etal-2023-imu2clip} and LLMSense~\cite{ouyang2024llmsense} rely on feeding long narrative logs to LLMs to provide the sensor information necessary for accurately answering the questions, making them vulnerable to LLMs’ well-known struggles with long-context reasoning~\cite{huang2023survey,gu2023mamba}. Others like DeepSQA~\cite{xing2021deepsqa} and OneLLM~\cite{han2024onellm} were built for short-duration data (i.e., less than one minute) and cannot scale effectively to long-duration, high-frequency inputs of multiple days with their current model and training pipeline.
In contrast, \Method overcomes these limitations by introducing a dedicated query stage that retrieves relevant sensor embeddings from a pre-encoded database. This design enables accurate answers to quantitative questions in \Dataset~\citesensorqa.}

\begin{figure}
\begin{center}
\vspace{-4mm}
\includegraphics[width=0.8\textwidth]{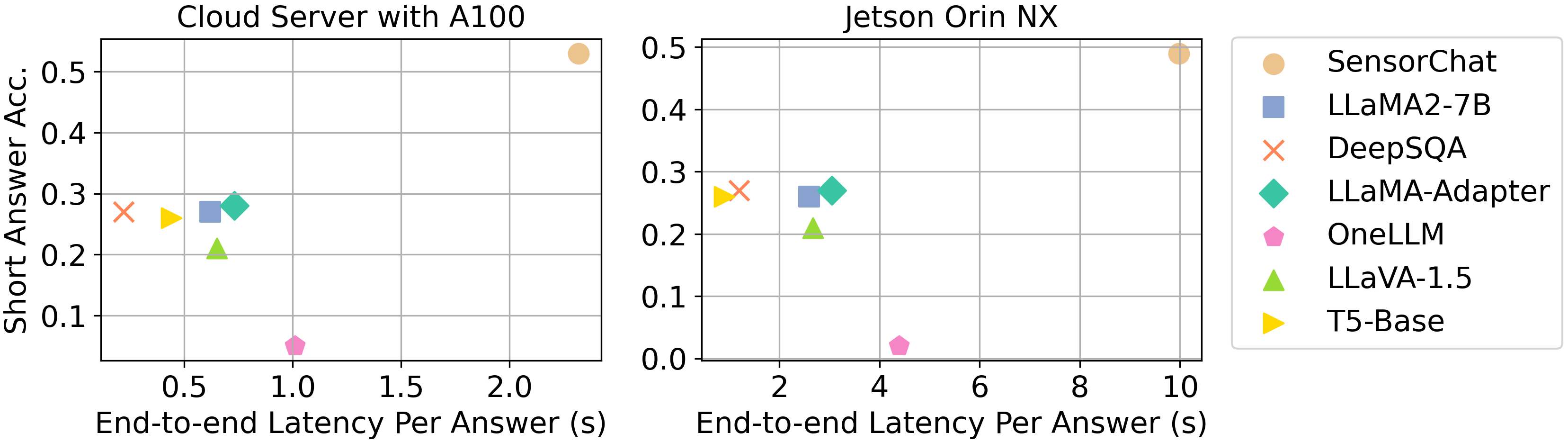} 
\vspace{-4mm}
\caption{The answer accuracy and latency trade-offs of all methods measured on the cloud and edge platforms. 
We evaluate \MethodC on A100~\cite{a100} and \MethodE on Jetson Orin NX~\cite{jetsonorin}.
All LLM-based models are quantized to 4-bit weights with AWQ~\cite{lin2023awq} for Jetson Orin deployments.}
\label{fig:latency}
\end{center}
\vspace{-4mm}
\end{figure}

\subsection{End-to-End Answer Generation Latency}
To evaluate efficiency, we measure the end-to-end answer generation latency of \MethodC and \MethodE on their respective platforms - cloud server and NVIDIA Jetson Orin. For a fair comparison, we quantize all LLM-based baselines to 4-bit weights using AWQ~\cite{lin2023awq} for the Jetson Orin experiments, matching \MethodE’s setup. \textbf{\Method takes an average generation latency of 2.3s on the cloud and 10.0s on the Jetson Orin}.  
Specifically, \MethodC takes an average of 1.2s for question decomposition, 0.5s for sensor data query, and 0.6s for answer assembly. \MethodE takes an average of 2.5s, 4.9s, and 2.5s for each stage, respectively.

The accuracy-latency trade-off of \Method and locally running baselines is shown in Fig.~\ref{fig:latency}. Despite higher latency due to its dual-LLM design, \Method outperforms other baselines in accuracy. \Method still achieves real-time responses on the cloud and can be deployed on resource-constrained devices with reasonable generation latency.
Notably, quantization causes negligible accuracy loss in \MethodE compared to the full-precision \MethodC. This is due to \Method's design, which converts sensor data into text, making the final answer assembly a purely language-based task where techniques like AWQ~\cite{lin2023awq} minimize accuracy degradation in language tasks after quantization.
We plan to further optimize the latency of \Method on edge devices via algorithm-system co-design in future work.
\section{Real User Study}
\label{sec:deployment}

\revise{Since the evaluations in Sec.\ref{sec:evaluation} focus on quantitative questions, we further conduct a user study to assess whether \Method can assist users in real-world scenarios. This real user study evaluates \Method's generalizability to personalized sensor setups and to arbitrary user-generated questions by having the participants collect their own sensor data and pose questions of personal interest. }
Our study is approved by the Institutional Review Board Committee.


\textbf{User Study Setup} We recruited eight volunteers (five males and three females) who were instructed to follow their normal daily routines while carrying a smartphone with the ExtraSensoryApp~\cite{vaizman2018extrasensory}. The smartphone models include Huawei Mate 10 Pro, LG G7 ThinQ and Google Pixel 2. The app automatically collected multimodal sensor data and transmitted it to \Method whenever a network connection was available. Reporting activity labels was optional for the participants. The data collection phase lasted one to three days, with valid samples ranging from 52 to 1366 minutes. Sensor data availability varied due to factors such as phone model and usage patterns. After the data collection phase, volunteers were invited to interact with \MethodC in-person through a chat-based graphical user interface (GUI). \revise{Details of the GUI are provided in Appendix~\ref{appendix:gui}.} They were encouraged to ask \textit{any} questions about their lives during the data collection phase and observe \MethodC's response generation in real time. In the final step, we gathered feedback from the participants including ratings on answer content, latency, and practical value of \Method on a scale from 1 to 5. 
We specifically select \MethodC for the user study to evaluate \Method's full functionality in real-world conditions. We leave user evaluation and optimization of \MethodE in a purely edge scenario for future work.

\begin{figure*}[t]
  \centering
  \includegraphics[width=0.98\textwidth]{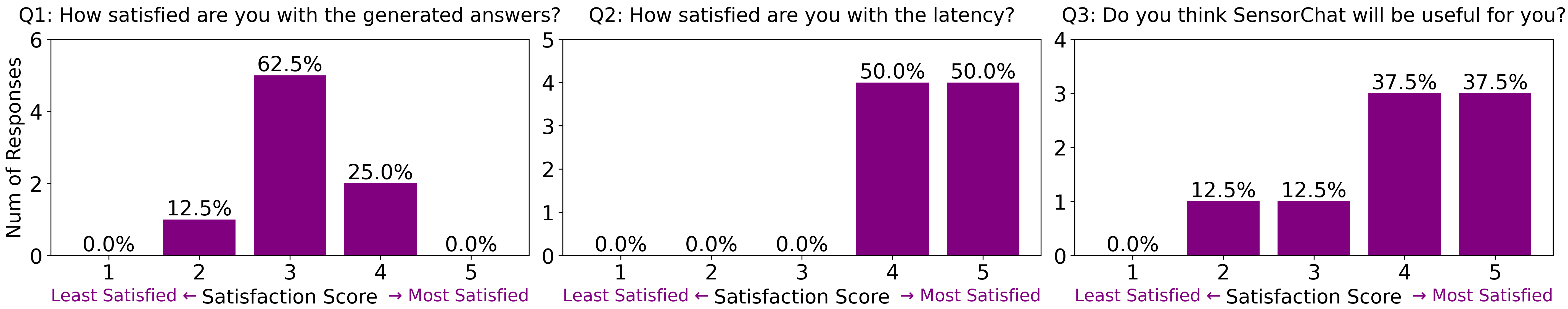}
  \vspace{-4mm}
  \caption{\revise{Feedback ratings from eight participants on answer satisfaction (left), latency satisfaction (middle), and the usefulness of \Method (right). Participants recognized \Method’s answer quality (on both quantitative and qualitative questions) along with its low latency and practical value.}}
  \vspace{-4mm}
  \label{fig:user_study}
\end{figure*}

\textbf{User Feedback Results}
Fig.~\ref{fig:user_study} shows the feedback ratings from eight participants. \textbf{\Method received an average score of 3.12 for answer content, 4.50 for latency, and 4.00 for practical utility}, highlighting its potential for real-world use.
\revise{We observed that users asked a wide range of questions, including both quantitative and qualitative ones. While some of the questions raised by participants overlap with the quantitative questions in \Dataset~\citesensorqa, many are qualitative in nature and are not covered by the dataset. The 3.12 average score suggests that \Method can handle both types of questions, which is a difficult task. More details of the questions asked by the users are presented in Appendix~\ref{appendix:user-study}.} 
Participants appreciated the natural tone of the responses, such as ``\textit{The answers were formatted in an easy to understand way}.'' Some concerns were noted about accuracy, including ``\textit{some numbers were a little off}'' and ``\textit{it mentioned activities I never did}.''
We believe these issues stem from the limited ability of the sensor encoder to generalize from the dataset to real-world users. In several cases, the sensor query stage produced noisy outputs and occasionally detected activities that did not occur.
Fortunately, generalizing models across different settings is a well-studied problem~\cite{xu2023practically}. Applying these techniques to \Method may improve its real-world performance, which we leave for future work.

In terms of latency, all participants give positive feedbacks regarding the end-to-end answer generation latency, with comments such as "\textit{I do not feel as if I had to wait a long time for the answers}" and "\textit{I think it is faster than I thought previously}".
Most participants are positive in terms of the practical utility of \Method in their daily lives, e.g., ``\textit{I believe it can help with my wellness management significantly.}'' The less positive comments mentioned the challenge of adapting \Method to individual users for generating more personalized and useful responses, an issue that we leave for future exploration.



\textbf{Generalization to Qualitative and Open-Ended Questions}
Participants have asked \revise{qualitative} and open-ended questions to \Method that do not exist in \Dataset, providing a more comprehensive assessment of \Method.
For example, one participant asked, ``\textit{How would you rate my lifestyle and what improvements do you suggest?}'' \Method responded with, ``\textit{I would rate this lifestyle as 2 out of 5 stars and would suggest you improve on your exercise and talking,}'' based on the duration of each activity where exercise and talking were lacking.
\revise{One advantage of \Method is that it can handle qualitative questions with the same three-stage pipeline. \Method effectively handles qualitative questions by comprehensively querying a list of contexts that are relevant to the qualitative question, such as "sitting", "exercising", and "talking" in the above example. \Method then passes the full set of context information to the answer assembly stage, where the LLM uses both this context and its world knowledge to reason about the user's lifestyle.}
In our study, participants appreciated the ability of \Method to effectively handle both quantitative and qualitative questions, e.g., "\textit{I think it could be very useful to be able to answer these qualitative and quantitative questions about my lifestyle,}" highlighting its broad applicability to diverse real-life scenarios.
\revise{We plan to conduct a larger-scale user study with a broader participant pool in the future to further investigate the generalizability of \Method across diverse age and gender groups.}

\section{Discussion \& Future Work}
\label{sec:future-work}


\textbf{Generalizability of \Method}
\revise{In this work, we design \Method address two main categories of questions: quantitative and qualitative. \Method provides accurate answers to quantitative questions by a successful fusion between LLMs and a sensor data query stage. The LLM interprets the question and generates natural language answers, while the query stage extracts relevant context from long-duration, high-frequency sensor data. Using the same pipeline, \Method also handles qualitative questions by passing richer context summaries to the LLM for high-level reasoning. However, \Method can still produce imprecise answers for time-related queries. Future work may explore more advanced LLMs, such as OpenAI’s o3-mini~\cite{openaio3} or DeepSeek~\cite{deepseek}, which perform better in numerical reasoning.}
\Method can also incorporate more sensors via adding additional modality encoders.
Additionally, \Method can serve a larger user base by collecting more data and equipping with better generalization designs such as~\cite{xu2023practically}. \Method is a general and flexible framework designed for fusing long-duration multimodal sensors knowledge with language and answering versatile questions.



\textbf{Evaluation Metrics} \Method uses traditional metrics in natural language processing and the precision of the answers in evaluation. While these metrics excel in objectiveness, they do not capture subjective user satisfaction, which is crucial for assessing a system's readiness for a broad market. 
We leave a comprehensive evaluation of \Method with subjective metrics as our future work.

\textbf{Efficiency of \Method} In this work, we demonstrate that \Method can be deployed on Jetson-level edge devices using common quantization techniques. However, further optimization is needed for real-time LLM service on edge devices. The long-term goal for systems like \Method is to run on mobile devices. Recent studies~\cite{liu2024mobilellm, zhuang2024litemoe} have explored enabling real-time LLM serving on mobile devices, and these techniques could be integrated into \Method in future work. Additionally, we recognize the potential synergy between \Method and vector databases~\cite{zhou2024llm} or retrieval-augmented generation~\cite{zhao2024retrieval}. The techniques in these domains can be integrated into \Method to improve query performance and efficiency.

\section{Conclusion}
\label{sec:conclusion}

Natural interactions between users and multimodal sensors can unlock the full potential of sensor data in real-world applications. However, existing systems struggle to process long-duration, high-frequency sensor data, often resulting in unsatisfactory answers to long-term questions.
In this paper, we present \Method, the first end-to-end system designed to handle long-duration, high-frequency time series sensor data for both qualitative and quantitative question answering. \Method introduces a novel three-stage pipeline including question decomposition, sensor data query and answer assembly.
All stages and interfaces are carefully designed to effectively fuse the natural language and the sensor modality.
Evaluation results and user studies demonstrate \Method's effectiveness in answering a wide range of qualitative and quantitative questions from state-of-the-art benchmarks and a real-world user study. The cloud version of \Method system supports real-time interactions, while the edge version is optimized to run on Jetson-level devices.

\begin{acks}
This work was supported in part by National Science Foundation under Grants \#2003279, \#1826967, \#2100237, \#2112167, \#1911095, \#2112665, \#2120019, \#2211386 and in part by PRISM and CoCoSys, centers in JUMP 2.0, a SRC program sponsored by DARPA.
\end{acks}
\bibliographystyle{ACM-Reference-Format}
\bibliography{sample-base}

\appendix
\section{Different Ways of Feeding Sensor Data to LLMs}
\label{appendix:input-format}

\revise{Existing works have mainly used four ways to feed sensor data to LLMs: (1) converting raw sensor data into text and feeding to LLMs~\cite{ji2024hargpt,xu2024penetrative,yang2024drhouse,hota2024evaluating,liu2023large,kim2024health,englhardt2024classification}, (2) visualizing raw sensor signals in images and feeding to LLMs~\cite{yoon2024my}, (3) processing raw sensor signals with a pretrained sensor encoder and feeding the encoded sensor embeddings to LLMs~\cite{moon2023anymal,han2024onellm,mo2024iot,chen2024sensor2text}, and (4) processing raw sensor signals with a pretrained sensor encoder that produces classification results in text, and then feeding the narrative text to LLMs~\cite{arakawa2024prism,moon-etal-2023-imu2clip,ouyang2024llmsense}. In the following lines, we explain why we choose (4) as the base framework for \Method, and why the first three approaches are not suitable for long-duration (more than one day), high-frequency (e.g., raw time series from IMU sensors) sensor data in our setting, as detailed in Sec.~\ref{sec:intro}.} 

\begin{enumerate}
    \item \revise{One common approach is to directly convert raw sensor measurements into text before passing them to LLMs, as seen in several recent works~\cite{ji2024hargpt,xu2024penetrative,yang2024drhouse,hota2024evaluating,liu2023large,kim2024health,englhardt2024classification}. However, we argue that this method is neither effective nor scalable for long-duration, high-frequency sensor data. To illustrate, consider using HARGPT~\cite{ji2024hargpt} with 40Hz IMU data. The prompt would need to encode time-series measurements like the following:}
    \begin{mytextbox2}
    \revise{\textbf{\#\#\# Instruction:} You are an expert of IMU-based human activity analysis. \\}
    \revise{\textbf{\#\#\# Question:} The IMU data is collected from smartphone and smartwatch with a sampling rate of 40Hz. The IMU data is given in the IMU coordinate frame. The three-axis accelerations and gyroscopes are given below. \\}
    \revise{Accelerations: \\}
    \revise{x-axis: 0.34979248, 0.34767151, ..., y-axis: -0.42715454, -0.42863464, ..., z-axis: -0.89616394, ... \\}
    \revise{Gyroscope: \\}
    \revise{x-axis: -0.11219972, -0.20297755, ..., y-axis: -0.08648433, -0.20838715, ..., z-axis: 0.03115442, ... \\}
    \revise{Please answer the following question based on the given information and IMU readings: \{Question\}. Please make an analysis step by step. \\}
    \revise{\textbf{\#\#\# Response:} }
    \end{mytextbox2}
    \revise{While this type of input is manageable for short durations (e.g., 10 seconds generating 2,400 values), it becomes infeasible for longer periods. For example, 24 hours of 6-axis IMU data at 40Hz would generate approximately 20 million values—equivalent to roughly 50 million tokens (assuming 10 characters per number and 1 token per 4 characters). This significantly exceeds the context limits of even the most advanced LLMs, such as GPT-4 (8K tokens)~\cite{gpt-4} or GPT-4o (128K tokens)~\cite{gpt-4}. In addition to input size limitations, LLMs are known to struggle with long-context reasoning~\cite{huang2023survey,gu2023mamba}, making it difficult to extract accurate answers from such lengthy inputs. While recent work has proposed extracting low-frequency features (e.g., activity labels, summary statistics) and converting them into text logs~\cite{xu2024autolife}, it remains an open question which features best preserve meaningful information across diverse user queries, while remaining within the token limits of current LLMs. For these reasons, we did not include baselines from this category. In contrast, \Method addresses these challenges by encoding long-duration raw sensor data into compact embeddings and incorporating a dedicated query stage, enabling scalable and accurate reasoning across extended time periods.}

    \item \revise{Visualizing raw sensor signals as images and feeding them into LLMs is another approach that may help models understand sensor data~\cite{yoon2024my}. However, the complexity and scale of the data in our scenario introduce challenges that typical visualization approaches are not well-equipped to address. Visualizing long-duration, high-frequency data often leads to dense and cluttered images, making it difficult to interpret or extract actionable insights. For instance, as shown in Fig.\ref{fig:signal_image}, even a simple plot of the first axis of smartphone IMU data for a single user over multiple days results in a figure that is hard to parse, especially when attempting to answer concrete time-related questions. Such time queries seek precise minute-level answers, which are common in QA benchmarks like \Dataset\citesensorqa.
    As such, we did not include direct comparisons with baselines in this category. While adaptively selecting relevant segments for visualization based on user queries is a potential direction, it remains an open challenge to determine appropriate time windows automatically. \Method avoids this issue by incorporating an explicit query stage to efficiently retrieve and reason over relevant sensor data.}

    \begin{figure*}[t]
      \centering
      \includegraphics[width=0.8\textwidth]{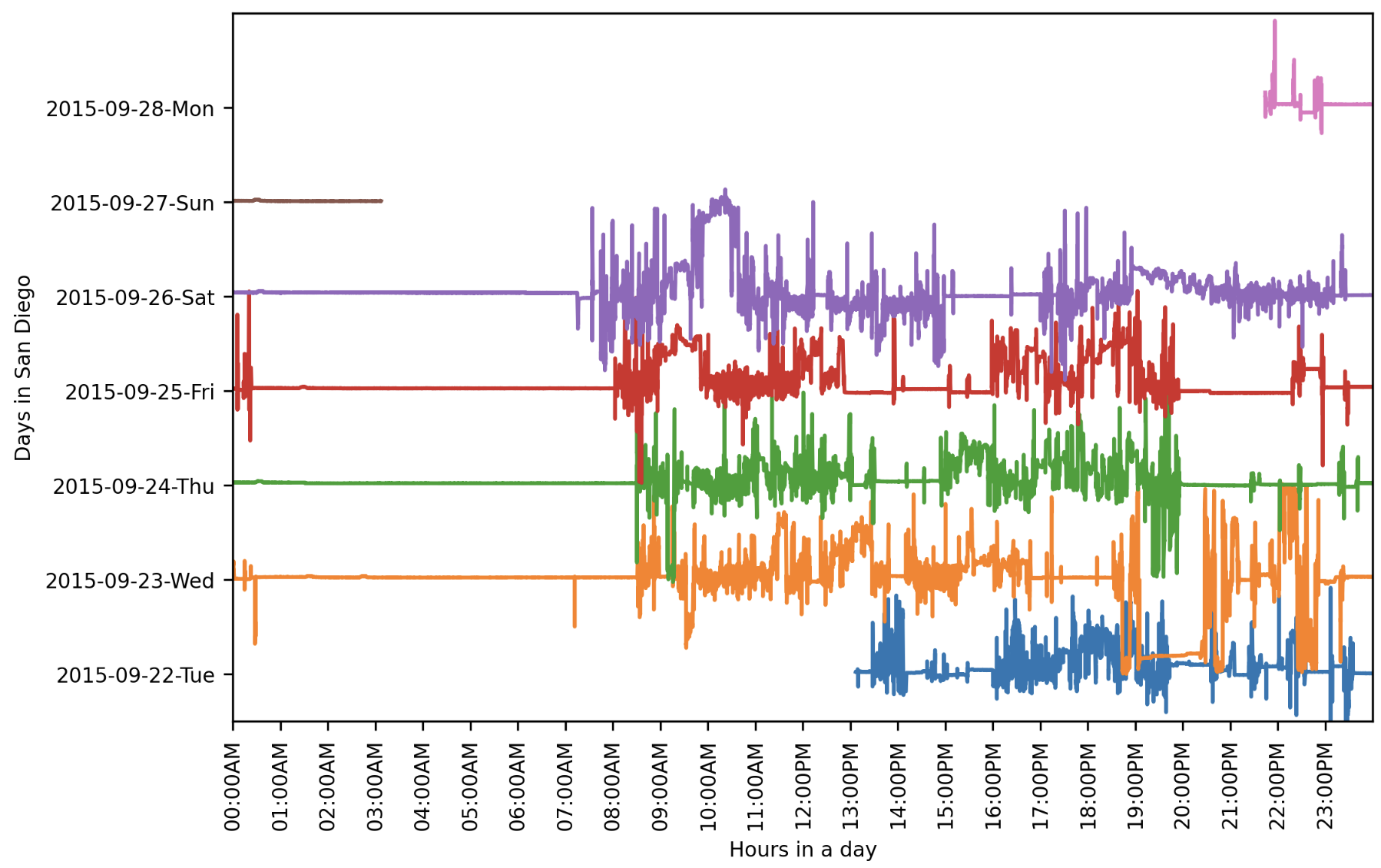}
      \vspace{-4mm}
      \caption{\revise{Visualization of the first axis of smartphone IMU signals. These visualizations pose extreme challenges in answering both quantitative and qualitative questions, such as ``\textit{How long was I at school?}'' or ``\textit{Did I go to school?}''}}
      \label{fig:signal_image}
      \vspace{-6mm}
    \end{figure*}
    
    \item \revise{Multimodal LLMs use a pretrained sensor encoder or adapter to compress raw sensor measurements into embeddings, which are then passed to an LLM~\cite{moon2023anymal,han2024onellm,mo2024iot,chen2024sensor2text}. In our evaluation (Sec.\ref{sec:evaluation}), methods like LLaMA-Adapter\cite{zhang2023llama}, LLaVA-1.5~\cite{liu2024improved}, and OneLLM~\cite{han2024onellm} fall into this category. These approaches perform well on tasks like video captioning or QA on short IMU sequences, but struggle to scale when applied to long-duration, high-frequency sensor data.
    For example, OneLLM converts every 10-second chunk of IMU data into a 512-dimensional embedding. Over a full day, this results in an embedding matrix of 8,640 rows and 512 columns. If each element were treated as a token, the total input size would reach 4.4 million tokens. Similar as case (1), such long input token sequence far exceeds the token limits of state-of-the-art LLMs and likely leads to inaccurate responses.}

    \item \revise{The last category of methods use neural network to associate the raw sensor signals with specific context labels, which are then concatenated into a narrative text to LLMs~\cite{arakawa2024prism,moon-etal-2023-imu2clip,ouyang2024llmsense}. In our evaluation (Sec.\ref{sec:evaluation}), IMU2CLIP+GPT4o~\cite{moon-etal-2023-imu2clip} and LLMSense~\cite{ouyang2024llmsense} are the latest contributions belonging to this category. These methods are well-suited to our setting, as they avoid passing long-duration sensor data directly to LLMs, addressing the interpretability and token limit issues seen in earlier approaches. Instead, they use LLMs for reasoning based on pre-processed text summaries. However, both IMU2CLIP+GPT-4o and LLMSense use a passive summarization approach similar to journaling systems~\cite{adiga2020daily,takahashi2013design}, by simply stitching together all timestamped activity descriptions into one long narrative. For example, the following shows the start of a typical narrative text generated by IMU2CLIP+GPT4o~\cite{moon-etal-2023-imu2clip}:}
    \begin{mytextbox}
    \revise{2015-09-30-Wed 13:17PM The person is at main workplace \\}
    \revise{2015-09-30-Wed 13:19PM The person is walking and talking at school and at main workplace while the phone is in hand \\}
    \revise{2015-09-30-Wed 13:20PM The person is sitting and doing computer work at school and at main workplace \\}
    \end{mytextbox}
    \revise{Such long, passively generated narrative text often includes irrelevant information and can exceed LLM token limits, leading to truncated inputs and missed details. To overcome this, \Method introduces a dedicated query stage that extracts only the sensor information relevant to the user’s question. This results in cleaner, more focused inputs for the final answer assembly stage, improving both clarity and accuracy of \Method's answers.}
\end{enumerate}

\revise{In summary, \Method adopts the strategy of (4) that is the most suitable for our scenario, that is, QA over long-duration, high-frequency raw sensor measurements. \Method further improves on the latest works by introducing a dedicated query mechanism to effectively process long-duration, high-frequency raw sensor signals while fusing knowledge with user questions.}

\section{Custom Query Function Design in \Method}
\label{sec:query-function}

In \Method, we design the following query functions to handle various real-life QA scenarios.
\begin{itemize}
    \item \texttt{CalculateDuration} is used for time comparisons, time queries, and existence questions.
    \item \texttt{DetectActivity} handles activity queries. \texttt{CountingFrequency} and \texttt{CountingDays} handle single- and multi-day counting. 
    \item \texttt{DetectFirstTime} and \texttt{DetectLastTime} are used for specific timestamp queries.
\end{itemize}
We emphasize that the function set can be expanded in the future to accommodate a broader range of queries.
However, since the \revise{query} function is selected by the LLM during question decomposition, it is crucial to limit the number of functions to avoid overwhelming or confusing the model.
Fortunately, one strength of \Method is its ability to transform questions into existing functions during question decomposition, eliminating the need to create new ones unnecessarily.
For instance, a question like "What did I do right after waking up?" can be decomposed into calls to \texttt{DetectFirstTime} and \texttt{DetectLastTime}, rather than requiring the creation of a new function such as \texttt{DetectNextActivity}.
To enable this transformation, we include a list of available functions in the prompt, as shown in Fig.~\ref{fig:question-decompose}.
This approach ensures that the current set of six \revise{query} functions is sufficient to effectively handle the vast majority of user queries.

\section{\Method Implementation Details}
\label{appendix:implementation-details}

The algorithm part of \Method is implemented with Python and PyTorch~\cite{paszke2019pytorch}.
For the offline sensor encoder, \Method uses a Transformer architecture~\cite{vaswani2017attention} with 6 encoder layers, 8 attention heads, and a feedforward network size of 2048 for time series data. For the low-frequency phone status data, \Method employs a fully connected layer as the encoder. The fusion layer is also implemented as a fully connected layer. The label encoder is initialized from the pretrained CLIP ViT-B/32 label encoder~\cite{radford2021learning}. Both the sensor and label embedding spaces share a dimension size of 512. Offline pretraining is performed with the partial-context loss proposed in Sec.~\ref{sec:pretraining} using a temperature scalar of $\tau=0.1$. We use the Adam optimizer with a learning rate of $1e-5$ over 100 epochs.

For question decomposition, we design two solution templates for each question category. Limiting the number of templates to two helps balance prompt length and performance.
For online data queries, we initialize the classifier $f$ as a multilayer perceptron with one hidden layer and a ReLU activation function. The hidden layer size is set to 512 and the query threshold is configured to $h = 0.5$.
For finetuning the LLM in answer assembly, we apply low rank adaptation (LoRA) finetuning~\cite{hu2021lora} on an NVIDIA A100 GPU, using a batch size of 8, a learning rate of 0.0002 and 10 epochs.
The LoRA rank $r$ is 16, the LoRA scaling factor \textit{alpha} is 16, and the LoRA dropout is 0.1.
During answer generation, \Method uses a max sequence length of 1024 and a generating temperature of $0.2$.
%

\section{Baselines Implementation Details}
\label{appendix:baseline-implementation-details}

We consider both closed-source and open-source baselines for a comprehensive analysis, including
\begin{itemize}[topsep=0pt, itemsep=0pt]
    \item \textbf{GPT-4~\cite{gpt-4}}, \textbf{T5~\cite{2020t5}} and \textbf{LLaMA~\cite{touvron2023llama}} are popular \textcolor{mygreen}{text-only} language models that only take questions as input.
    \item \textbf{GPT-4-Turbo~\cite{gpt-4} and GPT-4o~\cite{gpt-4}} are \textcolor{myred}{vision+text} baselines taking the activity graphs and the questions as inputs. For all \textcolor{myred}{vision+text} baselines, we feed the activity graphs in \Dataset~\citesensorqa, similar to Fig.~\ref{fig:example_qas}, along with the questions into the model.
    \item \textbf{LLaMA-Adapter~\cite{zhang2023llama}} is a recent \textcolor{myred}{vision+text} framework offering a lightweight method for fine-tuning instruction-following and multimodal LLaMA models. It integrates vision inputs (i.e., activity graphs from \Dataset~\citesensorqa) with LLMs using a transformer adapter. We utilize the latest LLaMA-Adapter V2 model.
    
    \item \textbf{LLaVA-1.5~\cite{liu2024improved}} represents state-of-the-art \textcolor{myred}{vision+text} model. LLaVA connects pre-trained CLIP ViT-L/14 visual encoder~\cite{radford2021learning} and large language model Vicuna~\cite{vicuna2023}, using a projection matrix. LLaVA-1.5~\cite{liu2024improved} achieves state-of-the-art performance on 11 benchmarks through simple modifications to the original LLaVA and the use of more extensive public datasets for finetuning.
    \item \textbf{IMU2CLIP+GPT-4o~\cite{moon-etal-2023-imu2clip}} is the state-of-the-art \textcolor{myblue}{sensor+text} method using IMU2CLIP~\cite{moon-etal-2023-imu2clip} to associate raw IMU signals with text and using \revise{GPT-4o} for answer generation. For each timestamp, IMU2CLIP searches for a text description that best matches the IMU data. The complete narrative text from all timestamps is then fed into \revise{GPT-4o}~\cite{gpt-4} with the question to generate an answer.
    \item \revise{\textbf{LLMSense~\cite{ouyang2024llmsense}} is the state-of-the-art model for high-level reasoning over long-term sensor data. LLMSense fuses \textcolor{myblue}{sensor+text} by converting neural network classification outputs into text, which forms a list of narrative text for all timestamps. This text is then fed to GPT-4o~\cite{gpt-4} for reasoning.}
    \item \textbf{DeepSQA~\cite{xing2021deepsqa}} trains a CNN-LSTM model with compositional attention to fuse \textcolor{myblue}{sensor+text} modalities and predict from a fixed and limited set of candidate answers given questions and IMU signals. We adapt their implementation to use the full-history timeseries data as input, to align with \Method's setup.
    
    \item \textbf{OneLLM~\cite{han2023onellm}} is a state-of-the-art multimodal LLM framework that processes \textcolor{myblue}{sensor+text} modalities using a universal pretrained CLIP encoder and a mixture of projection experts for modality alignment. We adapt their implementation and feed the full-history timeseries data from the IMU tokenizer.
\end{itemize}



We did not compare with the latest works of Sensor2Text~\cite{chen2024sensor2text}, PrISM-Q\&A~\cite{arakawa2024prism}, and DrHouse~\cite{yang2024drhouse} due to limited access to their open-source code and models.
\revise{Table~\ref{tbl:baseline-details} summarizes key implementation details for all methods evaluated on the SensorQA dataset~\cite{sensorqa}. In particular, we highlight important aspects of input data format and training methods.}

\begin{table}[!t]
\footnotesize
\centering
\caption{\revise{Comparison of baseline and \Method implementation. The baselines are adapted in various ways to handle the long-duration, high-frequency raw sensor data in our setting. The detailed input formats are adjusted based on their original input modality. For all methods, the expected output is a textual answer.}}
\vspace{-4mm}
\label{tbl:baseline-details}
\begin{tabular}{c|c|c|p{7em}|p{18em}} 
\toprule
\textbf{Method} & \textbf{Text Input} & \textbf{Image Input} & \textbf{Sensor Input} & \textbf{Training Method} \\ 
\midrule
GPT-4~\cite{gpt-4} & Question & - & - & Few-shot learning \\
T5-Base~\cite{2020t5} & Question & - & - & Full training \\
LLaMA2-7B~\cite{llama2} & Question & - & - & LoRA finetuning~\cite{hu2021lora} \\ \midrule
GPT-4-Turbo~\cite{gpt-4} & Question & SensorQA activity graphs & - & Few-shot learning \\
GPT-4o~\cite{gpt-4} & Question & SensorQA activity graphs & - & Few-shot learning \\
LLaMA-Adapter~\cite{zhang2023llama} & Question & SensorQA activity graphs & - & LoRA finetuning~\cite{hu2021lora} \\
LLaVA-1.5~\cite{liu2024improved} & Question & SensorQA activity graphs & - & LoRA finetuning~\cite{hu2021lora} \\ \midrule
IMU2CLIP+GPT4o~\cite{moon-etal-2023-imu2clip} & Question & - & Raw IMU signals & First train a neural network using IMU2CLIP~\cite{moon-etal-2023-imu2clip}. Then, apply few-shot learning on GPT4o~\cite{gpt-4} for reasoning on narrative text \\
LLMSense~\cite{ouyang2024llmsense} & Question & - & Raw multimodal sensor signals & First, train a neural network for context classification using cross-entropy loss. Then, apply few-shot learning with GPT-4o~\cite{gpt-4} for reasoning on narrative text \\ 
DeepSQA~\cite{xing2021deepsqa} & Question & - & Raw IMU signals & Full training \\ 
OneLLM~\cite{han2024onellm} & Question & - & Raw IMU signals & LoRA finetuning~\cite{hu2021lora} \\ \midrule
\textbf{\Method (This work)} & Question & - & Raw multimodal sensor signals & Use few-shot Learning in the question decomposition stage, and LoRA finetuning~\cite{hu2021lora} in the answer assembly stage. \\
\bottomrule
\end{tabular}
\vspace{-2mm}
\end{table}

\revise{\textbf{Input Format} We employ a wide range of state-of-the-art baselines that are designed for various modalities, including text, images or sensors. To ensure a fair comparison with \Method, we adapt the long-duration, high-frequency sensor data to align with the original modality format of each baseline. The detailed inputs for each baseline (covering text, image, and sensor modalities) are shown in Table~\ref{tbl:baseline-details}.}

\begin{itemize}
    \item \revise{For text-only models, i.e., \textbf{GPT-4~\cite{gpt-4}}, \textbf{T5-Base~\cite{2020t5}}, \textbf{LLaMA2-7B~\cite{llama2}}, we only feed the textual question into the model.}
    \item \revise{For models that allow image+text input, i.e., \textbf{GPT-4-Turbo~\cite{gpt-4}}, \textbf{GPT-4o~\cite{gpt-4}}, \textbf{LLaMA-Adapter~\cite{zhang2023llama}}, \textbf{LLaVA-1.5~\cite{liu2024improved}}, we feed the activity graphs from SensorQA~\cite{sensorqa} as the image input and the textual question as the text input. We did not adopt the approach of directly visualizing raw sensor signals in graphs and feeding them to LLMs~\cite{yoon2024my}. This is because such visualization of long-duration (more than one day), high-frequency sensor signals (such as those from IMU sensors) are hard to interpret and extract useful insights, as discussed in Appendix~\ref{appendix:input-format}. Instead, the activity graphs in SensorQA clearly illustrate a user's schedule over a multi-day period, making them more suitable for our QA scenario. Note that these baselines use oracle activity graphs as input, which include ground-truth activity information. However, even with access to ground-truth data, their QA performance remains significantly lower than that of \Method, which relies on a predictive model.}
    \item \revise{Both \textbf{IMU2CLIP+GPT-4o~\cite{moon-etal-2023-imu2clip}} and \textbf{LLMSense~\cite{ouyang2024llmsense}} rely on a pretrained sensor encoder to generate textual descriptions for each timestamp. The full sequence of narrative text is then fed into a LLM to perform high-level reasoning and produce the final answer. In IMU2CLIP+GPT-4o, we train the sensor encoder using the IMU2CLIP approach~\cite{moon-etal-2023-imu2clip}. The sensor encoder maps raw IMU signals into an embedding space and retrieves the most similar text embedding. For LLMSense, we train a neural network with the same architecture as the sensor encoder in \Method, using cross-entropy loss for context classification. The text description is constructed by combining all predicted positive context labels into a complete sentence, following a set of fixed language rules. Both baselines use GPT-4o~\cite{gpt-4} as the LLM backbone. We choose GPT-4o for its significantly larger context window (128K tokens vs. 8,192 in GPT-4) and better cost efficiency (compared to GPT-4-Turbo).}
    \item \revise{\textbf{DeepSQA~\cite{xing2021deepsqa}} and \textbf{OneLLM~\cite{han2024onellm}} are multimodal models that accept sensor+text input. We emphasize that both models take the complete history of raw IMU signals associated with a single user as input. DeepSQA uses separate neural networks to process raw IMU signals and textual questions respectively. OneLLM employs an IMU tokenizer to directly process raw IMU signals, which are then passed through a sequence of pretrained universal encoders and a trainable Mixture-of-Experts architecture. The resulting encoded sensor embeddings, along with the text prompt containing the question, are then input into LLaMA2-7B.}
\end{itemize}
\revise{For all baselines, we adopt the same hyperparameters and pretrained weights as released in their official codebases.}

\revise{\textbf{Training Method} As shown in Table~\ref{tbl:baseline-details}, all closed-source GPT-based models use a few-shot learning approach. 
This includes \textbf{GPT-4~\cite{gpt-4}}, \textbf{GPT-4-Turbo~\cite{gpt-4}}, \textbf{GPT-4o~\cite{gpt-4}}, \textbf{IMU2CLIP+GPT4o~\cite{moon-etal-2023-imu2clip}}, and \textbf{LLMSense~\cite{ouyang2024llmsense}}.
We evaluate both zero-shot and few-shot learning for these baselines, and find that the few-shot approach yields better answer quality and accuracy. 
To determine the number of examples to include in the prompts, we perform a grid search over ${2, 5, 10, 15}$ on a separate validation set and select 10 examples, which provide the best results.
The rest open-source models use full training or LoRA finetuning~\cite{hu2021lora}. More specifically, \textbf{T5-Base~\cite{2020t5}} and \textbf{DeepSQA~\cite{xing2021deepsqa}} are fully trained.
\textbf{LLaMA2-7B~\cite{llama2}}, \textbf{LLaMA-Adapter~\cite{zhang2023llama}}, \textbf{LLaVA-1.5~\cite{liu2024improved}}, \textbf{OneLLM~\cite{han2024onellm}} are finetuned with LoRA~\cite{hu2021lora}.}

\revise{\textbf{SensorChat vs. Baselines} We visualize \Method's setup in Table~\ref{tbl:baseline-details} for a clear comparison. Designed for long-term multimodal sensor interactions, \Method uses the user question and the full history of raw multimodal sensor signals as input. Unlike LLMSense~\cite{ouyang2024llmsense}, which converts classification results into a long narrative text and feeds it to an LLM for reasoning, \Method performs an explicit sensor data query in the second stage and passes only the relevant query results to the final LLM stage for answer generation. These query results are much shorter, with typically just a few sentences. As a result, \Method is more likely to generate accurate answers in the end. \Method's training procedure consists of two LLM-based stages, each employing a different training strategy based on the availability of high-quality datasets. The first question decomposition stage uses few-shot learning due to the lack of a high-quality dataset for this given task. The last answer assembly stage uses LoRA fine-tuning to fully leverage the SensorQA dataset~\cite{sensorqa}, allowing the model to produce answers that are both accurate and aligned with user-preferred styles.}

\section{Additional QA Examples}
\label{appendix:qualitative}

\begin{figure}[t]
    \centering
    \begin{subfigure}[b]{0.85\textwidth}
        \centering
        \includegraphics[width=\textwidth]{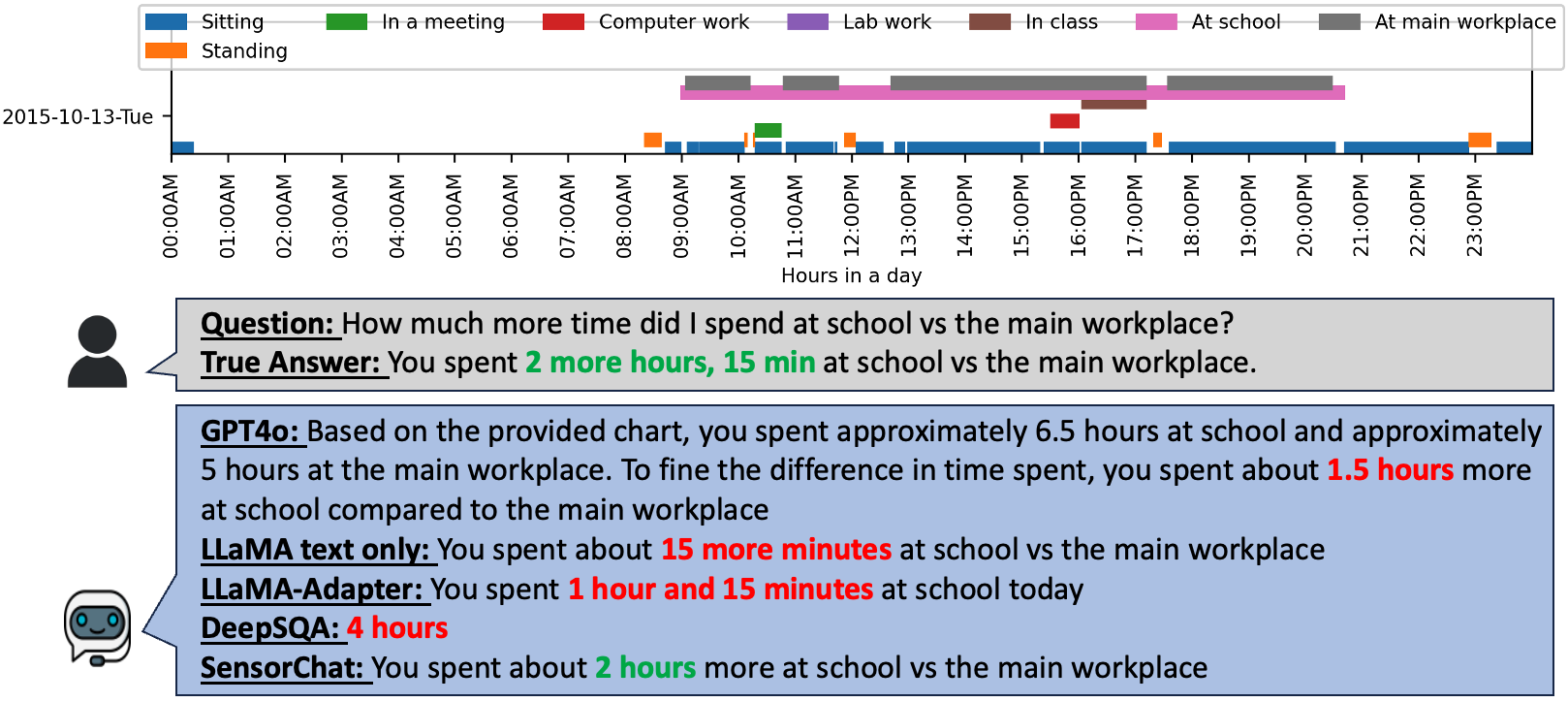}
        \vspace{-6mm}
        \caption{Example of a time query question on a single-day query.}
    \end{subfigure} 
    
    \begin{subfigure}[b]{0.85\textwidth}
        \centering
        \includegraphics[width=\textwidth]{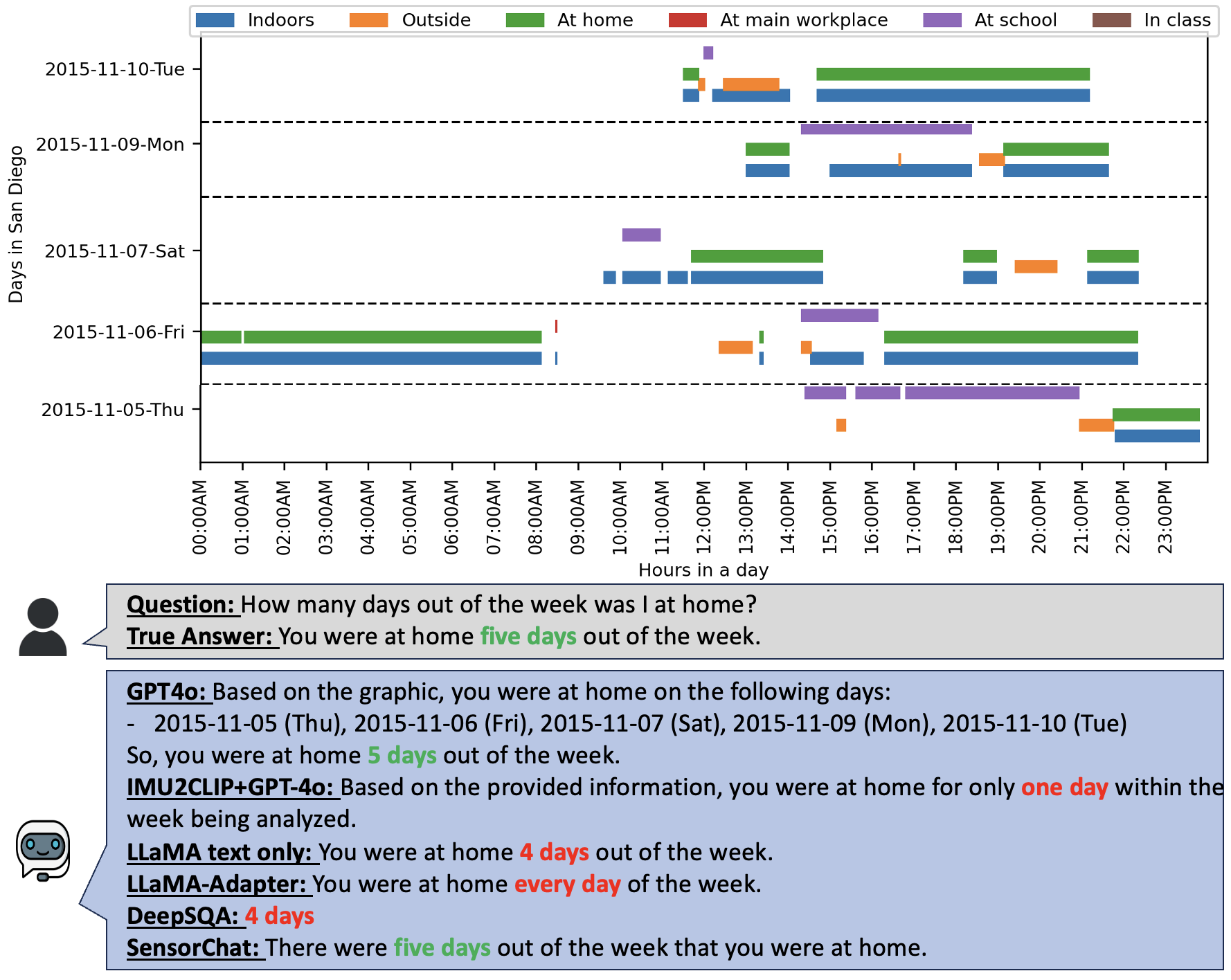}
        \vspace{-6mm}
        \caption{Example of a counting question on a multi-day query}
    \end{subfigure}
    \vspace{-4mm}
    \caption{More qualitative results of \Method in comparison to state-of-the-art methods. \revise{\Method delivers accurate answers for both single-day and multi-day queries, even for complex questions that require multi-step reasoning.}}
    \label{fig:qual_results_appendix}
    \vspace{-4mm}
\end{figure}

We illustrate a qualitative comparison of \Method with the top-performing baselines in Fig.~\ref{fig:qual_results_appendix}. 
Specifically, Fig.~\ref{fig:qual_results_appendix} (a) focuses on time-related queries and (b) focuses on counting questions, which are the most challenging for SOTA methods given the long-duration time series sensor inputs (see Sec.~\ref{sec:motivation}).
Key phrases in the answers are highlighted in the green if they closely match the true answer, and in the red if they do not. The two presented examples are very challenging for state-of-the-art baselines. 
\Method, on the other hand, consistently produces more accurate answers which can be attributed to its novel three-stage pipeline.

Answering the question in Fig.~\ref{fig:qual_results_appendix} (a) requires two major steps: (i) calculating the total time spent in school and in the main workplace, and (ii) computing the difference between them.
\Method accurately answers it by first decomposing and then querying the durations for ``at school'' and ``in the main workplace'' respectively, resulting in ``\textit{You spent 11 hours and 27 minutes at school Tuesday. You spent 9 hours and 3 minutes at main workplace}''. Finally, \Method integrates the above text and determines the time difference during the answering stage.
Although \Method's answer is 15 minutes off from the true value, possibly due to inaccuracies in the sensor encoder or LLM reasoning, \Method still approximates the ground truth with an accuracy unmatched by other baselines.

In Fig.~\ref{fig:qual_results_appendix} (b), counting the total days spent at home requires long-term reasoning where  LLaMA-Adapter~\cite{zhang2023llama} and DeepSQA~\cite{xing2021deepsqa} usually fall short.
The performance of IMU2CLIP+GPT-4o~\cite{moon-etal-2023-imu2clip} is constrained by GPT-4o’s token limit. When the narrative log generated from raw sensor data exceeds this limit, the overflow is discarded, leading to errors in reasoning. In this example, GPT-4o~\cite{gpt-4} is the only baseline that correctly answers the count-based question by leveraging the activity graph, which clearly indicates the user's presence at home each day and allows for simple counting. However, GPT-4o may struggle with more complex questions, such as those requiring precise duration calculations.
In contrast, \Method decomposes this question and queries the duration of "at home" on "each day", then leaving the counting task to the answering stage.
The collaboration across the three stages allows \Method to effectively manage a wide range of tasks, particularly those requiring multi-step reasoning and quantitative analyzes, which highlights \Method's advancements over existing works.

\section{Ablation Studies}
\label{sec:ablation}

In this section, we comprehensively examine the impact of key design choices in \Method. Without loss of generality, we use \MethodC as the base model.

\begin{table}[!t]
\footnotesize
\centering
\caption{Impact of three major stage in \MethodC. Bold values highlight the best results. \revise{The full \Method with all three stages achieve the best results, highlighting the indispensable contribution from each stage.}}
\vspace{-4mm}
\label{tbl:ablation}
\begin{tabular}{c|ccc|c} 
\toprule
\textbf{Setup} & \multicolumn{3}{c|}{\textbf{Full Answers}} & \textbf{Short Answers} \\ 
& Rouge-1 ($\uparrow$) & Rouge-2 ($\uparrow$) & Rouge-L ($\uparrow$) & Accuracy ($\uparrow$) \\
\midrule
w/o Question Decomposition & 0.73 & 0.57 & 0.71 & 0.35 \\
w/o Sensor Data Query & 0.72 & 0.62 & 0.72 & 0.26 \\ 
w/o Answer Assembly & 0.26 & 0.08 & 0.24 & 0.0 \\
\midrule
Full \Method & \textbf{0.77}& \textbf{0.62} & \textbf{0.75} & \textbf{0.54} \\
\bottomrule
\end{tabular}
\vspace{-2mm}
\end{table}

\textbf{Impact of Each Stage in \Method} 
We first evaluate the individual contribution of each stage in \Method by removing one of them from the pipeline. 
By removing question decomposition, we use a fixed and general decomposition templates for all questions. Removing sensor data query reverts the model to a language-only approach. By removing answer assembly, we directly output the queried sensor context from the second stage.
Table~\ref{tbl:ablation} summarizes the results on full and short answers including both quality and quantitative accuracy.
As observed, removing any stage leads to a significant performance drop. 
In \Method, all three stages must work collaboratively to deliver high-quality, accurate answers across diverse question types in \Dataset.
Among the three stages, the answer assembly stage has the most significant impact, as it directly influences the final output. Removing it results in severely degraded performance, with near-zero accuracy on short answers. However, the question decomposition and sensor data query stages are equally crucial.

\begin{table}[!t]
\footnotesize
\centering
\caption{Impact of various designs for sensor-text pretraining in \MethodC. Bold values highlight the best results.}
\vspace{-4mm}
\label{tbl:ablation-sensor-feature}
\begin{tabular}{c|c|c|c} 
\toprule
\small
\textbf{Sensor Data} & \textbf{Training Loss} & \textbf{Online Querying} & \textbf{Multiple Choices} \\ 
& & Accuracy ($\uparrow$) & Answer Accuracy ($\uparrow$) \\
\midrule
Statistical features & Partial-Context Loss & 0.91 & 0.62 \\
Time series & IMU2CLIP~\cite{moon-etal-2023-imu2clip} & 0.90 & 0.61  \\ \midrule
Time series & Partial-Context Loss & \textbf{0.98} & \textbf{0.70} \\
\bottomrule
\end{tabular}
\vspace{-2mm}
\end{table}


\textbf{Impact of Sensor Features and Pretraining Loss Functions}
We next evaluate the impact of sensor features and loss functions during offline encoder pretraining.
Specifically, we compare using statistical features (e.g., mean acceleration) versus raw time series inputs, and IMU2CLIP loss~\cite{moon-etal-2023-imu2clip} versus our proposed contrastive sensor-text pretraining loss for partial context (see Sec.\ref{sec:pretraining}). 
For IMU2CLIP, text samples are generated by combining all valid labels into one sentence. We report the online querying accuracy and multiple-choice answer accuracy to assess the influence to sensor information extraction.
As shown in Table~\ref{tbl:ablation-sensor-feature}, statistical features result in low querying and answer accuracies. This validates our motivation to design \Method that high-frequency time series sensor data are critical for fine-grained activity information. IMU2CLIP training, even with fine-grained data, yields poorer querying and answering accuracies, highlighting its limited ability associating sensor embeddings from partial text query. Our proposed loss function, which aligns sensor and text encoders for partial context queries, proves more effective. These findings emphasize the importance of selecting appropriate sensor features and loss functions during pretraining in order to achieve high-performance QA.

\textbf{Impact of LLM Design Choices}
We finally evaluate the impact of various design choices for LLMs. 
In question decomposition, we assess the contribution of in-context learning (ICL), chain-of-thought (CoT) techniques, and different backbone LLMs.
The results are summarized in Table~\ref{tbl:ablation-detailed-design}. 
Both ICL and CoT are crucial for high-quality and accurate answers. This is because an effective question decomposition improves sensor data queries. The solution templates in ICL are more essential to \Method as removing ICL reduces answer accuracy by 11\%. CoT enhances reasoning and slightly boosts accuracy by 1-5\%. Interestingly, using a more advanced backbone (GPT-4 vs. GPT-3.5-Turbo) results in minimal improvement in answer quality, as GPT-4, while generating richer text, does not follow instructions as well as GPT-3.5-Turbo according to our observation.

For answer assembly, we evaluate the effectiveness of finetuning compared to zero-shot or few-shot learning, as well as different LLaMA backbones. As shown in Table~\ref{tbl:ablation-detailed-design}, zero-shot learning results in poor performance, while few-shot learning improves answer quality but still lags behind finetuning. This highlights that finetuning is the most effective approach when a dataset like \Dataset is available. Using a more advanced LLaMA backbone, such as LLaMA3-8B, has minimal impact. Finetuning and the dataset prove to be more important than the model architecture during answer assembly.



\begin{table}[!t]
\footnotesize
\centering
\caption{Impact of various design choices for LLMs in \MethodC. Bold values highlight the best results. \revise{The results highlight the necessity of in-context learning (ICL) and chain-of-thought (CoT) during question decomposition, and using finetuning instead of zero- or few-shot learning during answer assembly.}}
\vspace{-4mm}
\label{tbl:ablation-detailed-design}
\begin{tabular}{c|c|c|ccc|c} 
\toprule
\small
\textbf{Stage} & \textbf{Setup} & \textbf{Model in Stage} & \multicolumn{3}{c|}{\textbf{Full Answers}} & \textbf{Short Answer} \\ 
& & & Rouge-1 ($\uparrow$) & Rouge-2 ($\uparrow$) & Rouge-L ($\uparrow$) & Accuracy ($\uparrow$) \\
\midrule
& w/o ICL & GPT-3.5-Turbo & 0.75 & 0.59 & 0.72 & 0.43 \\
Question & w/o CoT & GPT-3.5-Turbo & 0.76 & 0.61 & 0.74 & 0.50 \\ 
Decomposition & Full & GPT-4 & \textbf{0.77} & \textbf{0.62} & \textbf{0.75} & 0.49 \\
& Full & GPT-3.5-Turbo & \textbf{0.77} & \textbf{0.62} & \textbf{0.75} & \textbf{0.54} \\ \hline
& Zero-shot learning & LLaMA2-7B & 0.16 & 0.07 & 0.14 & 0.0 \\
Answer & Few-shot learning & LLaMA2-7B & 0.43 & 0.29 & 0.41 & 0.24 \\
Assembly & Finetuning & LLaMA3-8B & 0.76 & 0.61 & 0.74 & 0.53 \\
& Finetuning & LLaMA2-7B & \textbf{0.77} & \textbf{0.62} & \textbf{0.75} & \textbf{0.54} \\
\bottomrule
\end{tabular}
\vspace{-2mm}
\end{table}

\section{Sensitivity Analyses}
\label{sec:sensitivity}

\begin{figure}[t]
   \centering
    \setlength{\tabcolsep}{0.2pt}
\begin{tabular}{cccc}
        \vspace{-2mm}
        \includegraphics[width=0.22\textwidth, height=2.2cm]{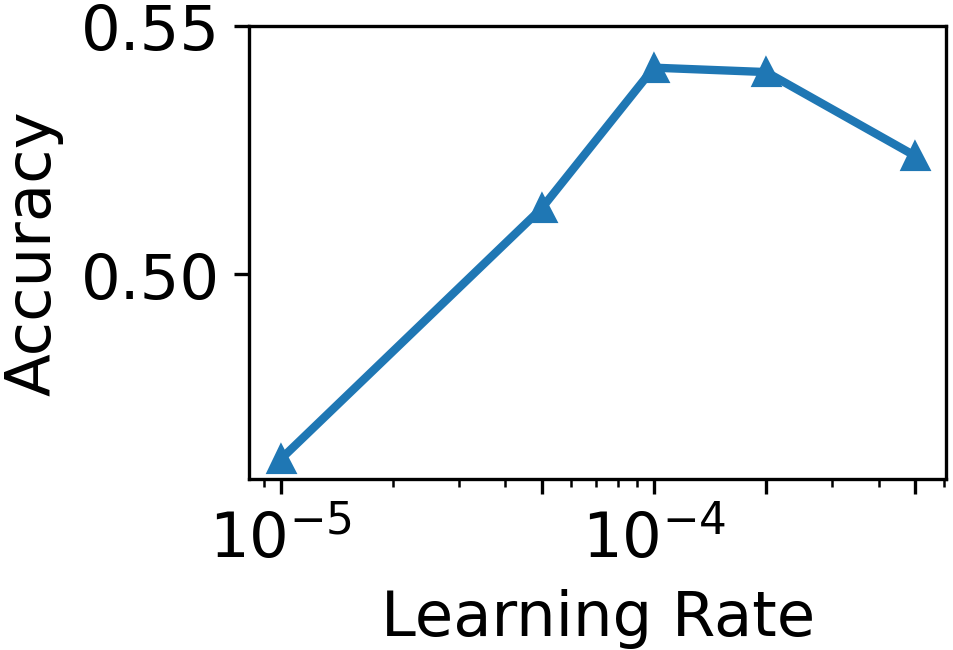} &
        \includegraphics[width=0.22\textwidth, height=2.15cm]{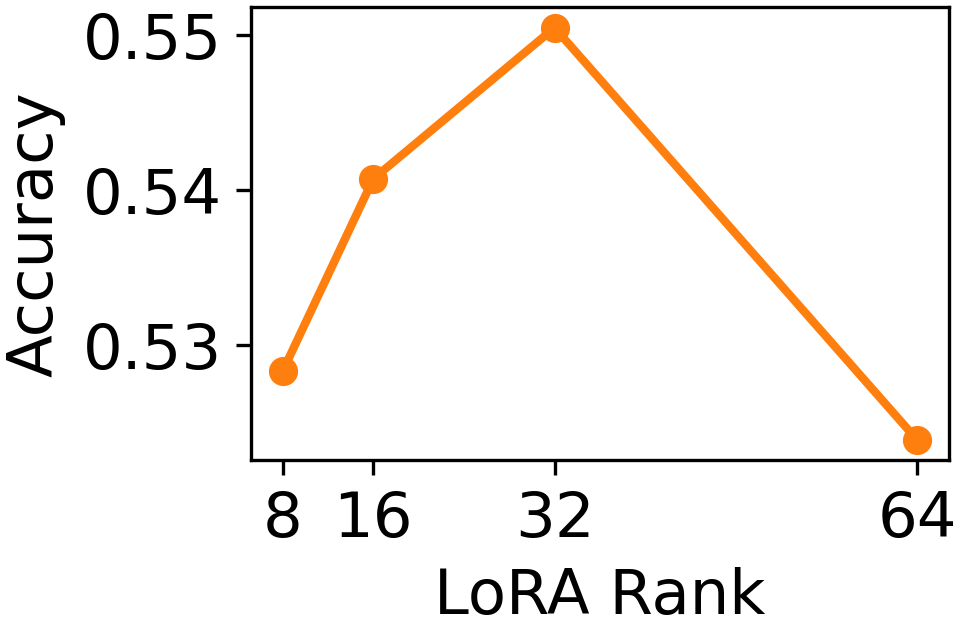} &
        \includegraphics[width=0.22\textwidth, height=2.2cm]{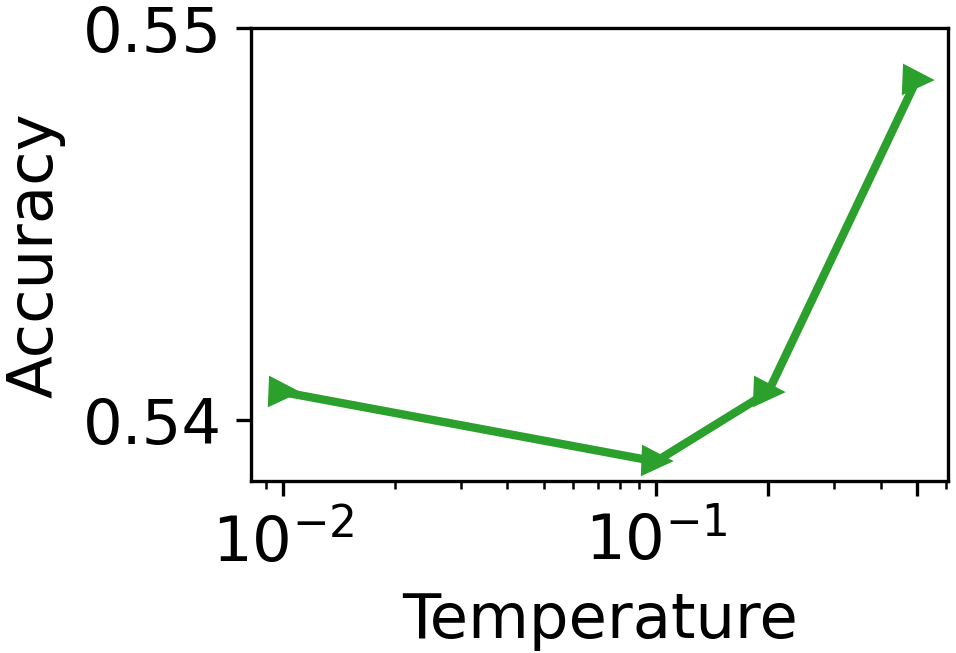} &
        \includegraphics[width=0.22\textwidth, height=2.15cm]{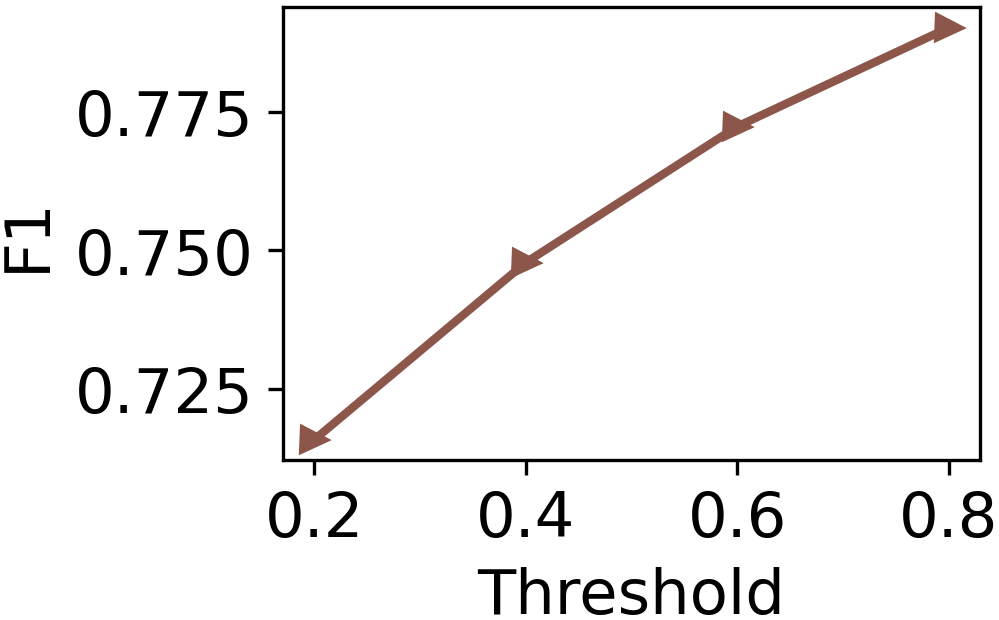} \\ 
\end{tabular}
\vspace{-3mm}
    \caption{Sensitivity of key hyperparameters \revise{including learning rate (leftmost), rank in LoRA finetuning (middle left), LLM generating temperature (middle right) and query threshold in \Method (right most).}}
    \label{fig:sensitivity}
    \vspace{-3mm}
\end{figure}

Fig.~\ref{fig:sensitivity} shows the sensitivity of key parameters in \Method.
while the less sensitive ones are omitted due to space limitation.
The default parameter setting is the same as described in Sec.~\ref{sec:system-implementation}.
We mainly focus on evaluating the short answers to assess the parameters' impact on factual information extraction.

\textbf{Learning Rate in Answer Assembly} 
Fig.~\ref{fig:sensitivity} (leftmost) shows the short answers accuracy for learning rates of $\{1e-5, 5e-5, 1e-4, 2e-4, 5e-4\}$ during finetuning.
Larger learning rates result in bigger gradient steps during LoRA finetuning, with $1e-4$ providing the best performance for our task.

\textbf{LoRA Rank in Answer Assembly}
Fig.~\ref{fig:sensitivity} (middle left) shows the short answers accuracy for LoRA ranks of $\{8, 16, 32, 64\}$ during finetuning.
The rank affects the size of the LoRA adapter weights. Higher ranks mean more parameters and a larger weight space to optimize. 
For our task, varying ranks have little impact on final accuracy, with rank 32 achieving the best performance for short answers.

\textbf{Generating Temperature in Answer Assembly}
Fig.~\ref{fig:sensitivity} (middle right) shows the short answers accuracy for generating temperatures of $\{0.01, 0.1, 0.2, 0.5\}$ during the final answer generation.
Higher temperatures instruct the LLM to use more ``creativity''. For our task, varying temperatures have negligible impact on the short answers accuracy, indicating minimal impact to the sensor information extraction.

\textbf{Query Threshold in Sensor Data Query} Fig.~\ref{fig:sensitivity} (rightmost) shows the F1 scores of online querying at different query thresholds $h=\{0.2, 0.4, 0.5, 0.6, 0.8\}$. 
Raising the threshold $h$ excludes less confident positive predictions, which can improve F1 scores. However, this may also overlook some detailed events, potentially reducing answer accuracy. Ideally, $h$ should be calibrated individually for each user to achieve the best results.

\section{Impact of Sensor Data Duration}
\begin{figure}[t]
  \centering
  \includegraphics[width=0.36\textwidth]{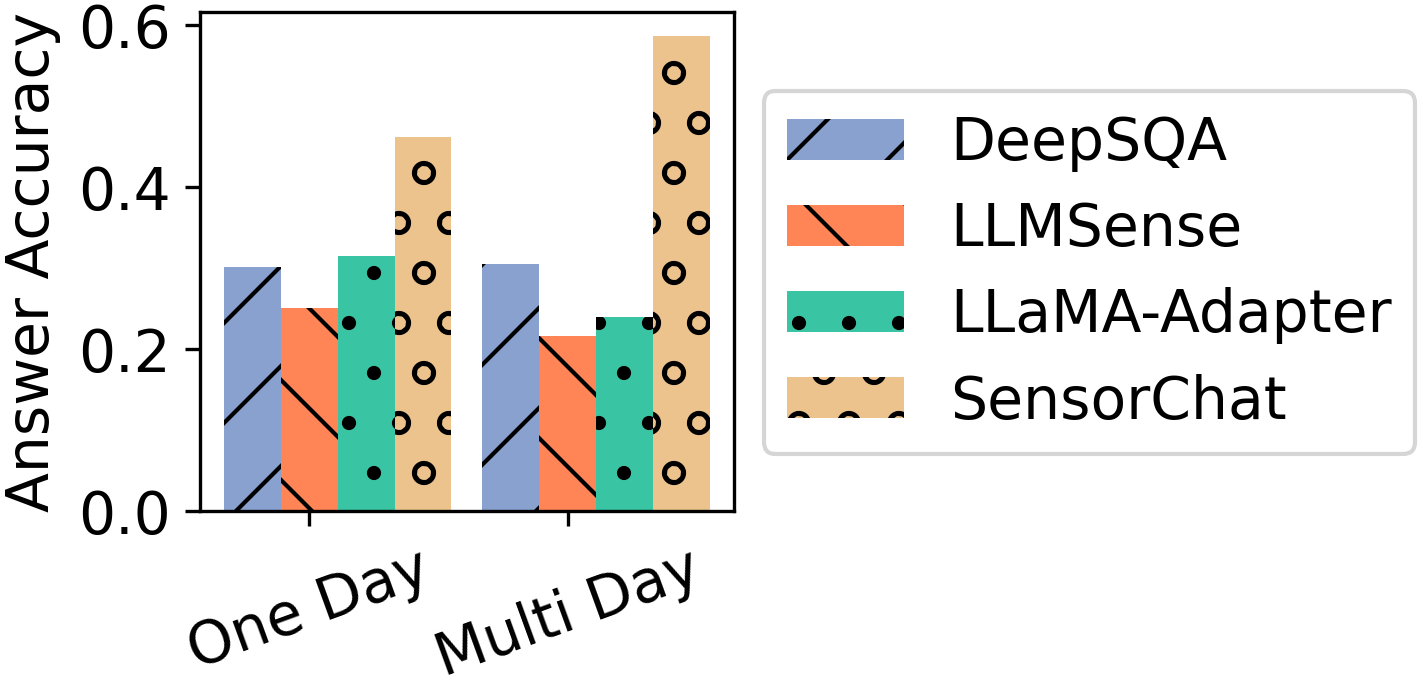}
  \vspace{-4mm}
  \caption{\revise{Performance of \Method compared to representative baselines on \Dataset~\citesensorqa, categorized by the duration of sensor data associated with each user question.}}
  \vspace{-4mm}
  \label{fig:duration}
\end{figure}
\revise{Fig.\ref{fig:duration} presents the performance of \Method and representative baselines across different sensor data durations using \Dataset\citesensorqa. We categorize user questions based on the associated sensor data into single-day and multi-day groups and report short answer accuracy.
The baselines include DeepSQA~\cite{xing2021deepsqa}, LLMSense~\cite{ouyang2024llmsense}, and LLaMA-Adapter~\cite{zhang2023llama}, each representing a mainstream approach. As discussed in Sec.~\ref{sec:qa-performance}, these methods struggle with long-duration, high-frequency sensor data. Fig.\ref{fig:duration} further shows that their performance declines when scaling from single-day to multi-day inputs: LLaMA-Adapter drops from 0.31 to 0.23, and LLMSense from 0.25 to 0.21. This highlights their limited ability to reason over long-term sensor data. Specifically, LLaMA-Adapter compresses sensor signals into embeddings before passing them to the LLM, which becomes less effective as the time span grows. LLMSense’s drop in performance stems from LLMs’ known difficulty with long-context reasoning~\cite{huang2023survey,gu2023mamba}, often resulting in confusion and inaccurate responses.
In contrast, \Method achieves higher accuracy on multi-day queries (0.58) than on single-day ones (0.46), demonstrating its strength in handling long-duration questions. However, this does not mean \Method underperforms on short durations. The slightly lower single-day accuracy is due to the nature of those queries, which demand precise time-based answers more frequently than multi-day queries, such as, “\textit{How long did I exercise today?}” As discussed in Sec.~\ref{sec:qa-performance}, short answer accuracy is a stricter metric for such questions, leading to lower scores compared to multi-day queries (e.g., “\textit{On which day did I exercise?}”).}

\section{Generalization and Robustness}
\label{sec:generalizability}
\begin{figure}[t]
  \begin{subfigure}[b]{0.55\textwidth}
        \centering
        \includegraphics[width=\textwidth]{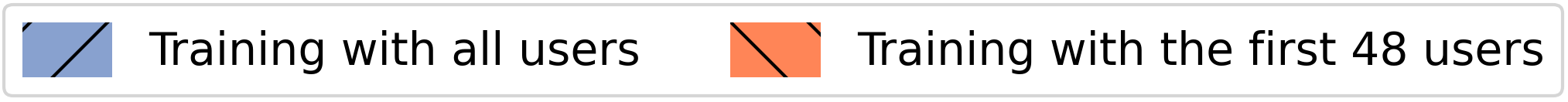}
        \vspace{-5mm}
    \end{subfigure}

    \begin{subfigure}[b]{0.35\textwidth}
        \centering
        \includegraphics[width=\textwidth]{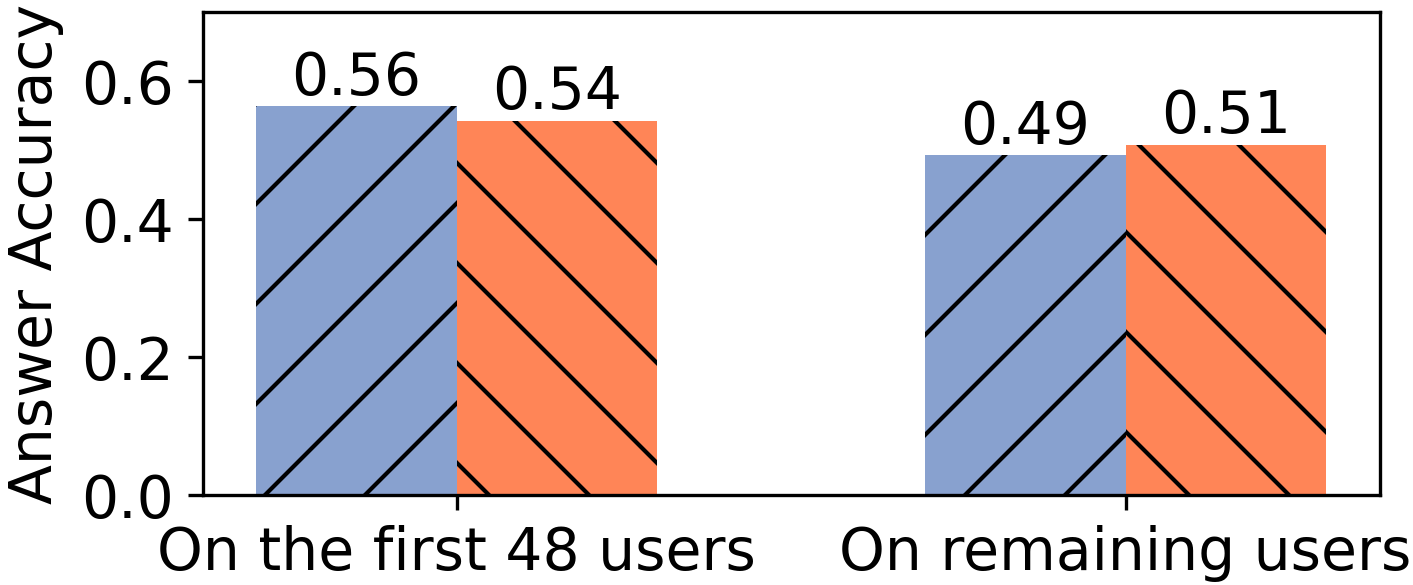}
        \vspace{-6mm}
        \caption{Average short-answer accuracy on different users.}
        \label{fig:exact_split}
    \end{subfigure} \hspace{0.02\textwidth} 
    \begin{subfigure}[b]{0.35\textwidth}
        \centering
        \includegraphics[width=\textwidth]{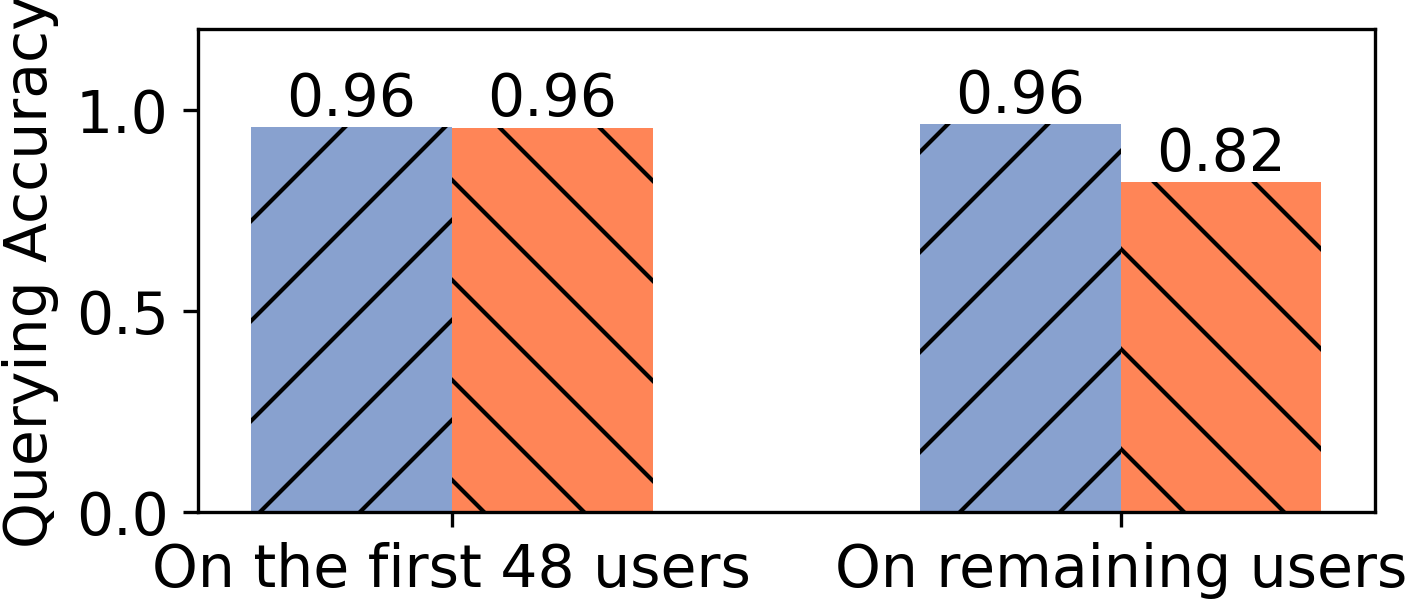}
        \vspace{-6mm}
        \caption{Average online querying accuracy on different users during sensor data query.}
        \label{fig:acc_split}
    \end{subfigure} 
    \vspace{-4mm}
    \caption{\revise{Generalization performance of \Method’s answer assembly stage (a) and online query stage (b) when trained on data from different users. Results show that the answer assembly stage generalizes well across users, while the sensor data query stage is more sensitive to shifts in user data distribution.}}
    \label{fig:generalization}
    \vspace{-4mm}
\end{figure}

We evaluate \Method's generalization and robustness across different users' sensor data and QA inputs. In addition to the standard 80/20 random split, we compare results with a split where training is performed on the first 48 out of 60 users and testing includes all users. To ensure a fair comparison, we equalize the training set size in both splits by duplicating samples in the smaller set.
Our evaluation focuses on two key learning processes in \Method: LLM fine-tuning in answer assembly and sensor-text encoder pretraining.

Fig.~\ref{fig:exact_split} presents the short-answer accuracy when fine-tuning on all users' QA data versus only the first 48 users. The results show similar accuracy for both seen and unseen users, demonstrating \Method's strong generalizability in answer assembly. This is likely due to \Method's design of treating answer assembly as a pure language task. Since all users' language tokens follow similar distributions in a sensor-based QA task, generalization remains robust across user variations.

Fig.\ref{fig:acc_split} presents the online querying accuracy when pretraining with all users' sensor data versus only the first 48 users. Unlike language fine-tuning, limiting sensor data to the first 48 users leads to accuracy degradation on unseen users due to variations in data distributions. Therefore, improving \Method's generalization to new users primarily depends on developing a robust sensor and text encoder, which is a well-studied problem in existing literature~\cite{xu2023practically}. We leave the investigation for combining with these techniques in future work.

\revise{\textbf{Generalizability Across Sensor Device Placements} We acknowledge that sensor data from vastly different body positions could introduce ambiguity in the encoder. However, detailed device placement information may not always be available, depending on the sensor setup.
In this work, \Method adopts a generalized approach without assuming prior knowledge of sensor placement. 
The key idea that \Method could work in this case is that, by training on a large-scale dataset collected from diverse users with varying device placements, the model learns a general similarity association between sensor data and context labels. When the user queries about a specific context of their lives, \Method could retrieve all sensor recordings sufficiently similar using the using learned similarities.
Additionally, if device placement information is available, we foresee that integrating \Method with recent sensor context augmentation techniques, such as~\cite{chowdhury2025zerohar}, could further enhance performance. We leave this as an avenue for future exploration.}

\section{Graphical User Interface of \Method}
\label{appendix:gui}

\begin{figure*}[t]
  \centering
  \includegraphics[width=0.95\textwidth]{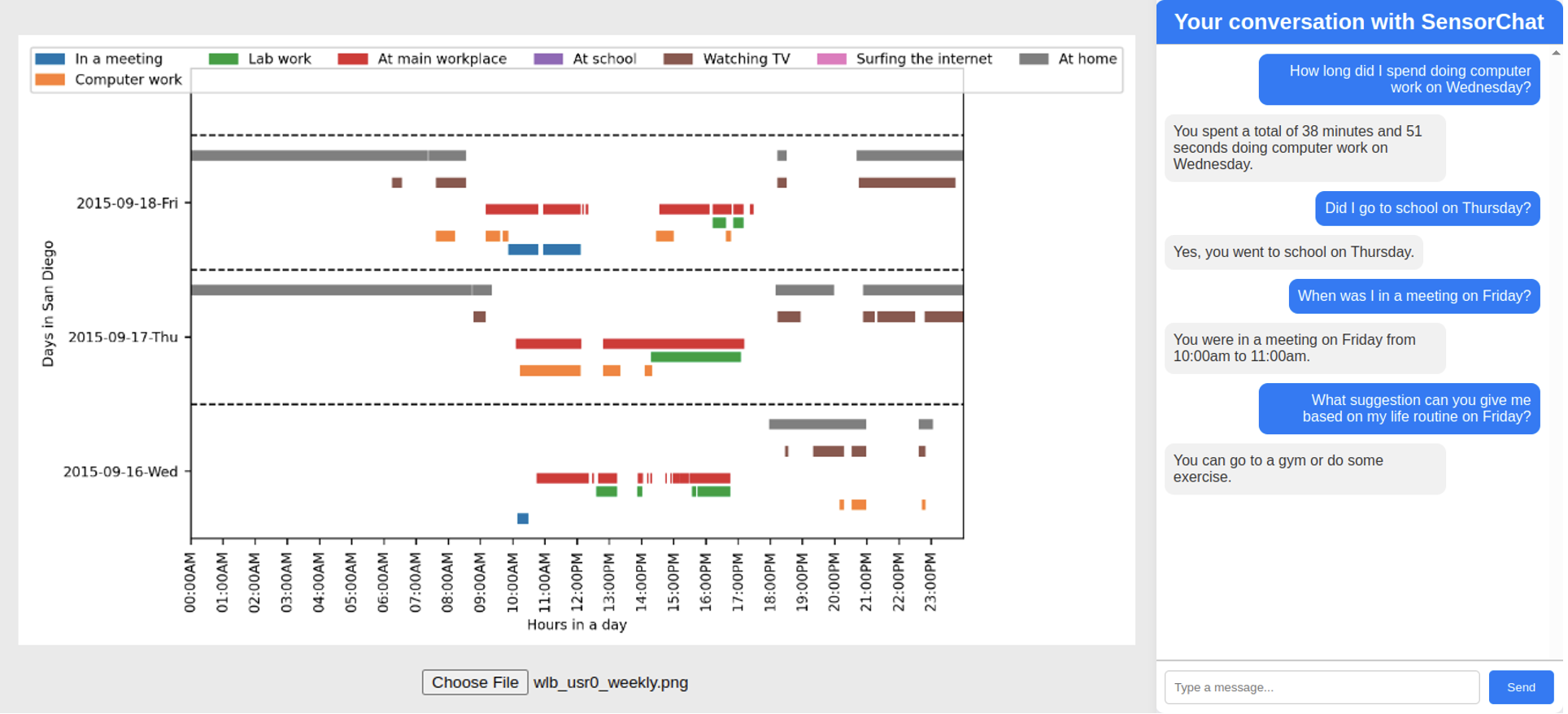}
  \vspace{-3mm}
  \caption{\revise{A prototype Graphical User Interface (GUI) for \Method. The left panel displays an optional activity visualization based on model predictions from raw sensor data. We plan to enhance interactivity by allowing users to select specific days and context labels of interest in the future. The right panel shows the main conversation between users and the \Method system: user questions appear in blue box on the right, and \Method responses appear in gray box on the left.}}
  \vspace{-4mm}
  \label{fig:gui}
\end{figure*}

\revise{Figure~\ref{fig:gui} presents a prototype of the \Method GUI used in our user study. Users primarily interact with \Method by posing questions and receiving responses, as illustrated in the right panel of the figure. These questions may be either quantitative or qualitative. In this example, the system operates with a low temperature setting for answer generation (see Appendix~\ref{sec:sensitivity} for additional sensitivity experiments), resulting in concise responses that emphasize key facts. This temperature parameter can be adjusted to better align with individual user preferences.
The left panel of the GUI includes an optional activity visualizer that provides users with an overview of their schedule. This visualizer can help remind users of their schedule and guide them toward more insightful questions. For lightweight deployments, such as the \MethodE mobile app on edge devices, this visualizer can be disabled, retaining only the conversation panel for streamlined interaction.}

\revise{We emphasize that this interface represents an initial prototype of \Method. Future enhancements will focus on improving interactivity, such as enabling users to select specific dates and context labels for visualization in the left panel. Additionally, we plan to incorporate a raw sensor data viewer for users interested in inspecting signals captured from their mobile or wearable devices.}

\section{Questions Asked in The Real User Study}
\label{appendix:user-study}

\revise{The goal of the real user study is to evaluate \Method's generalizability and practicality in real-world scenarios, extending beyond the predefined scope of \Dataset~\citesensorqa. We notice that some participant questions overlap with those in \Dataset, particularly when they are quantitative. The participants also raised a variety of qualitative questions that reflect their personal interests and are not included in \Dataset. After interacting with \Method and posing a mix of both qualitative and quantitative questions, participants reported fairly high satisfaction, demonstrating \Method's adaptability and usefulness in real-life scenarios.}

\revise{For example, the following quantitative questions are raised by the participants, which are similar to those in \Dataset:}

\begin{mytextbox}

\revise{How much time did I spend sitting down vs standing up?}

\revise{How much time did I spend doing lab work?}

\revise{How much time on average did I spend on entertainment for the past 2 days?}

\revise{Did I perform any household chores, if so, what shores did I do?}

\revise{What did I do after having lunch?}

\end{mytextbox}

\revise{The following qualitative questions are posed by the participants during real user study, which are \textbf{not} present in \Dataset:}

\begin{mytextbox}

\revise{What do you suggest to make me more active?}

\revise{How would you rate this lifestyle and what do you suggest that I improve on?}

\revise{Did I exercise enough for a young adult?}

\revise{Do I like cycling?}

\revise{Am I attentive when talking to my co-workers?}

\end{mytextbox}

\end{document}